\documentclass[11pt]{article}

\usepackage[final]{acl}

\usepackage{times}
\usepackage{latexsym}

\usepackage[T1]{fontenc}

\usepackage[utf8]{inputenc}

\usepackage{microtype}

\usepackage{inconsolata}

\usepackage{graphicx}

\usepackage{times}
\usepackage{latexsym}
\usepackage[T1]{fontenc}
\usepackage[utf8]{inputenc}
\usepackage{microtype}
\usepackage{inconsolata}
\usepackage{amsmath}
\usepackage{amssymb}
\usepackage{booktabs}
\usepackage{graphicx} 
\usepackage{multirow}
\usepackage{tabularx}
\usepackage{makecell}

\title{Integrated and Cross-Architecture Interpretation of LLM Reasoning}

\author{
  \textbf{Leonardo Matthew Yauw\textsuperscript{1, 2}\footnotemark[\value{footnote}]}\thanks{Equal contribution} \qquad
  \textbf{Wei-Bin Kou\textsuperscript{1}\footnotemark[\value{footnote}]} \qquad
  \textbf{Yujiu Yang\textsuperscript{1}\thanks{Corresponding author}}
\\
  \textsuperscript{1}Tsinghua Shenzhen International Graduate School,
  \textsuperscript{2} Harbin Institute of Technology, Shenzhen
\\
  \small{
    \href{mailto:leonardo.matthew.yauw@gmail.com}{leonardo.matthew.yauw@gmail.com},
    \href{mailto:wbkouqvb@sz.tsinghua.edu.cn}{wbkouqvb@sz.tsinghua.edu.cn},
    \href{mailto:yujiu.yang@sz.tsinghua.edu.cn}{yujiu.yang@sz.tsinghua.edu.cn}
  }
}

\begin{document}
\maketitle
\begin{abstract}
Understanding how LLMs reason is hindered by a practical asymmetry: while their generated outputs are observable, the underlying reasoning patterns remain opaque. Relying on single probes, such as Mutual Information Peak (MIP) or Deep-Thinking Ratio (DTR), risks underestimating the genuine inferential structure. To respond this deficiency, we present an Integrated, cross-Architecture Reasoning (IAR) framework, designed to provide a unified approach to LLM reasoning interpretability. Specifically, we first propose to use bandwidth-calibrated MIP coupled with Tukey IQR peak-detection to isolate reasoning-crucial tokens at the output layer. Second, we perform an overlap analysis between MIP-picked tokens and DTR-deep tokens to trace the cross-layer trajectories of those tokens. This also discloses whether reasoning-crucial tokens are computation-intensive as well, further facilitating to understand how reasoning patterns evolve across model layers. Finally, we apply a Jaccard stability metric over multi-domain problems to verify if the MIP-identified tokens are reasoning quality-guaranteed. Extensive experiments on six models across four domains (mathematics, code, logic, and common sense) demonstrate IAR's generalizable interpretation capabilities across architectures. Our code is available at \href{https://github.com/LeonardoMatthew/IAR}{https://github.com/LeonardoMatthew/IAR}.
\end{abstract}

\section{Introduction}

Large reasoning models such as DeepSeek-R1 \citep{deepseek2025} and OpenAI's o1 \citep{openai2024o1} have demonstrated remarkable capabilities in complex problem-solving through chain-of-thought (CoT) reasoning \citep{sprague2025cot,stechly2024chain}. Building on the original CoT paradigm, subsequent works have explored decoding strategies that aggregate multiple reasoning paths \citep{lei2025learning} and search over branching solution trees \citep{liao2025tpo}. In addition, recent evidence indicates that scaling test-time compute can outperform scaling model parameters \citep{chen2026rethinking}. By generating explicit intermediate steps before committing to a final answer, these methods achieve strong performance on challenging mathematical, coding, and scientific benchmarks. However, this success is hindered by a fundamental methodological asymmetry: while the generated reasoning chains are readily observable \citep{ranaldi2024aligning}, the internal computational trajectories that produce them remain opaque. In particular, it is unclear which tokens carry genuine computational significance and which merely reflect surface-level fluency in a reasoning chain.

\begin{figure*}[t]
\includegraphics[width=\linewidth]{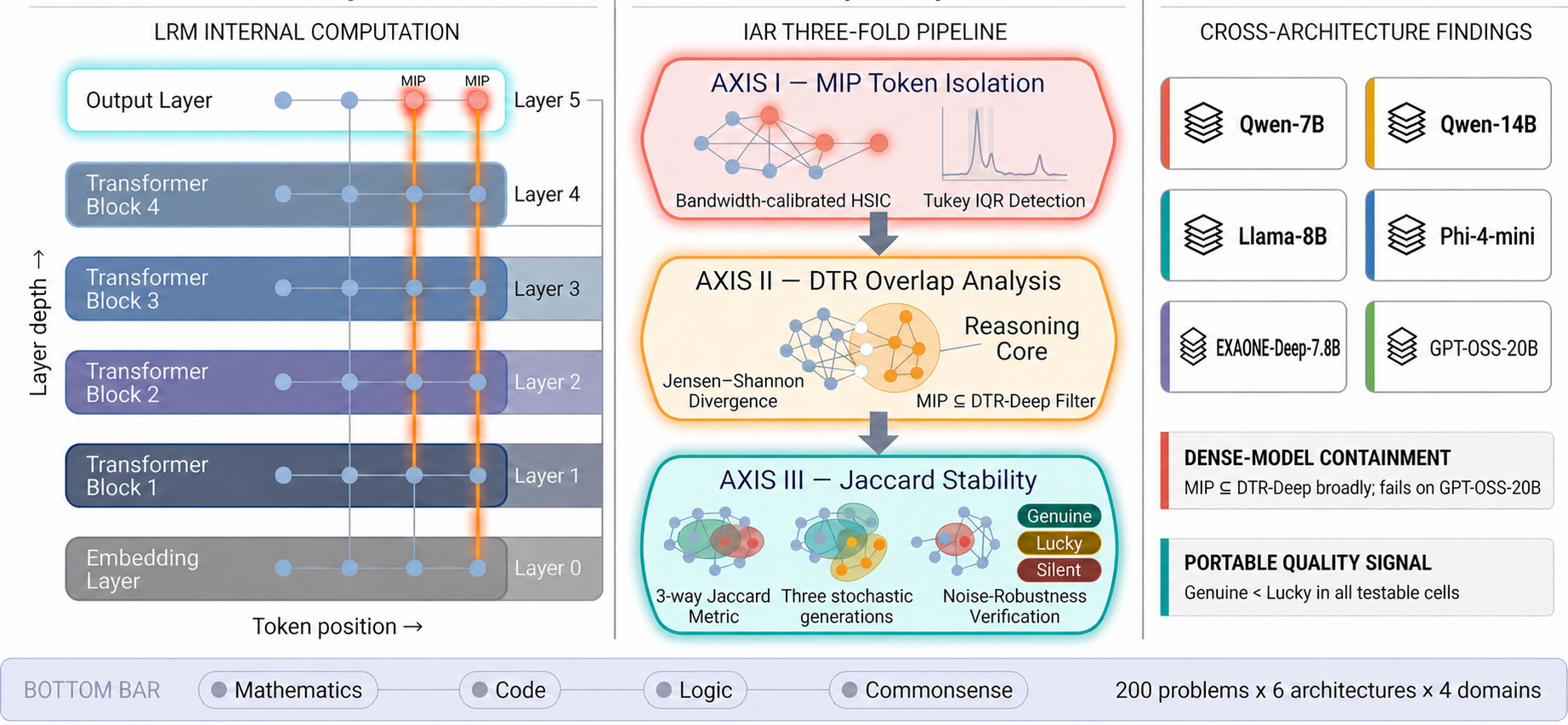}
\vspace{-0.66cm}
\caption{Overview of the proposed IAR framework.}
\label{Fig:IAR_overview}
\vspace{-0.38cm}
\end{figure*}

Recent works have begun to probe this opacity from multiple perspectives. \citet{qian2025} discovered that certain tokens exhibit sudden spikes in mutual information (MI) with the gold answer (termed MI peaks (MIP)), and showed that these MIPs correspond to reflective ``thinking tokens'' critical for reasoning performance, thereby characterizing \emph{reasoning-crucial} tokens at the output layer. \citet{chen2026} introduced the Deep-Thinking Ratio (DTR), which measures per-token \emph{computational depth} and \emph{cross-layer trajectories} via computing the similarity between each hidden layer and the final layer. 

Despite these achievements, two gaps remain. First, existing MIP and DTR interpretation probes work in isolation and disclose the LLM reasoning mechanism partially. Concretely,  MIP localizes tokens whose output representations carry strong information about the gold answer, but says nothing about the depth of computation invested in producing them. The DTR quantifies per-token computational depth through inter-layer divergence, but fires on too many tokens to serve as a discriminative interpretability signal on its own. Used in isolation, either probe risks underestimating genuine inferential structure, i.e., MIP ignoring the layered computation that gives rise to a reasoning-relevant token, DTR conflating deep computation with answer-relevance. Second, prior analyses have been confined to a \emph{single architectural family}. Without cross-architecture evaluation, it is impossible to distinguish real interpretation signals that reflect fundamental reasoning properties from those that are artifacts of a specific model.

To fill both gaps, we propose the \textbf{Integrated, Architecture-agnostic Reasoning interpretation paradigm (IAR)}, a unified framework that jointly reveals reasoning along three integral axes through a three-fold pipeline. First, we \emph{isolate reasoning-crucial tokens} at the output layer via bandwidth-calibrated MIP coupled with a Tukey IQR peak-detection rule, where architecture-wise bandwidth calibration is introduced to prevent kernel collapse under the HSIC estimator \citep{gretton2005}. Second, we \emph{compute DTR-deep tokens independently} via inter-layer Jensen--Shannon divergence, and perform an overlap analysis between MIP tokens and DTR-deep tokens. This helps to trace back MIPs' cross-layer trajectories and establishes whether MIP acts as an \emph{interpretable filter} on DTR, identifying the sparse, reasoning-relevant subset of computationally deep tokens. Third, we \emph{verify noise-robustness and reasoning quality} through a three-way Jaccard stability metric computed based on MIP token set across stochastic generations. We leverage this stability metric to operationalize the problem-level Genuine/Lucky/Silent taxonomy of \citet{sahoo2026}, allowing us to rigorously test whether aggregate MIP statistics can discriminate genuine reasoning from lucky guesses and silent failures across architectures.
The proposed IAR interpretation framework is illustrated in Fig.~\ref{Fig:IAR_overview}.

We instantiate IAR on six models (Qwen-7B, Qwen-14B, Llama-8B, Phi-4-mini, EXAONE-Deep-7.8B, and GPT-OSS-20B) across four domains (mathematics, code, logic, commonsense). Comprehensive experiments and extensive analyses show that (i) MIP-identified reasoning-crucial tokens and DTR-deep tokens are correlated, indicating that reasoning-crucial tokens are computation-intensive as well; (ii) this containment pattern generalizes across model architectures, offering a transferable reasoning structure in LLMs; and (iii) the reasoning quality discrimination via MIP statistics is portable across architectures.

In summary, the main contributions of this paper are highlighted as follows:
\begin{itemize}
    \vspace{-0.3cm}
    \item We propose the IAR Reasoning interpretation paradigm, a unified framework that jointly reveals LLM reasoning along three integral axes: \emph{reasoning-relevance, computation depth, and noise-robustness}.
    \vspace{-0.3cm}
    \item IAR contains a three-fold pipeline that (i) isolates reasoning-crucial tokens via bandwidth-calibrated MIPs at output layer; (ii) traces back MIP tokens' cross-layer trajectories and to test if these reasoning-crucial tokens are also computation-intensive; and (iii) verifies IAR's noise-robustness of the identified MIP tokens and their cross-layer trajectories.
    \vspace{-0.3cm}
    \item We conduct comprehensive evaluations on six models across four domains (mathematics, code, logic, commonsense) to demonstrate the cross-architecture interpretation capabilities of LLM reasoning.
\end{itemize}

\section{Related Work}
\label{sec:related}

\subsection{Reasoning in Large Language Models}
Large models have wide range of applications \cite{liu2025application,kou2025pfedlvm,yang2026route,wang2025comprehensive,kou2025label,chen2026mgfn++}. Chain-of-thought (CoT) prompting \citep{wei2022chain,kojima2022,zhang2026pixelcraft} has
established that eliciting intermediate reasoning steps substantially
improves performance on mathematical, coding, and commonsense tasks. Subsequent work extended this paradigm along two directions: aggregating multiple sampled chains via self-consistency \citep{wang2023selfconsistency,chia2024reasoning} and searching over branching reasoning structures such as Tree-of-Thoughts \citep{inoue2026wider,yao2023tot}. More recently, scaling test-time compute has been shown to rival or surpass scaling model parameters \citep{snell2025scaling,chen2026rethinking,geiping2026scaling}, and large reasoning models such as DeepSeek-R1 \citep{deepseek2025} and OpenAI o1 \citep{openai2024o1} have internalized extended CoT behavior through reinforcement learning. 

\subsection{Token-Level Interpretability via Mutual Information (MI)}
A first line of interpretability research characterizes reasoning at the \emph{output layer} \citep{sharma2024truth,hu2025efficient,xu2023eqmotion}. \citet{qian2025} observed that a small subset of tokens exhibits sharp spikes in mutual information (MI) with the gold answer, and showed that these \emph{MI peaks} (MIP) correspond to reflective ``thinking tokens'' that are significantly important for final-answer correctness \citep{chen2024learning,yong2026think}. While these methods successfully identify \emph{where} reasoning-relevant information concentrates in the token sequence, they operate on output-layer representations alone and therefore provide no account of the layered computation that produces those tokens. Our work retains MIP as the reasoning-crucial probe but calibrates its architecture-wise to avoid kernel collapse under the Hilbert--Schmidt Independence Criterion (HSIC) estimator \citep{gretton2005}, and composes it with a cross-layer probe rather than using it in isolation.

\subsection{Cross-Layer Trajectory Analysis} Some works characterize reasoning at the \emph{computational depth} axis by examining how hidden representations evolve across layers \citep{li2026understanding,wang2024grokking}. Logit-lens \citep{phukan2025beyond} and tuned-lens \citep{belrose2023} decode intermediate layers back to the vocabulary space, revealing that predictions often stabilize only in the upper layers. \citet{chen2026} formalize this observation via the Deep-Thinking Ratio (DTR), which measures per-token computational depth as the similarity trajectory between each hidden layer and the final layer, and declares a token DTR-deep when its layer-wise predictive distribution settles into the final distribution only in the upper layers. DTR captures genuine cross-layer computation, but fires on a large fraction of tokens and is therefore weak as a discriminative interpretability signal on its own. In contrast to prior work that uses either output-layer MI or cross-layer depth alone, we treat the two probes as \emph{compositional}: MIP acts as an interpretable filter that selects which of the many DTR-deep tokens are also reasoning-relevant.

\subsection{Reasoning-Quality Assessment}
A third line of work assesses whether observed reasoning behavior actually reflects reasoning quality or merely stochastic success. Self-consistency \citep{wang2023selfconsistency,wang2024soft,taubenfeld2025confidence,min2024beyond} and related sampling-based diagnostics use agreement across sampled chains as a quality proxy. More recently, \citet{sahoo2026} introduced a behavioral partition of reasoning problems into \emph{Genuine}, \emph{Lucky}, and \emph{Silent} categories based on per-problem correctness patterns combined with an activation-consistency confidence score. Unlike token-level attribution, this partition is defined at the problem level and provides a quality signal independent of any single probe. We adapt this taxonomy as an \emph{external} behavioral ground truth, replacing Sahoo's activation-consistency score with a three-way Jaccard stability metric computed over MIP sets across stochastic generations. This substitution preserves the original behavioral semantics while making the partition usable for evaluating whether aggregate MIP statistics discriminate reasoning-quality regimes.

\begin{figure*}[t]
\includegraphics[width=\linewidth]{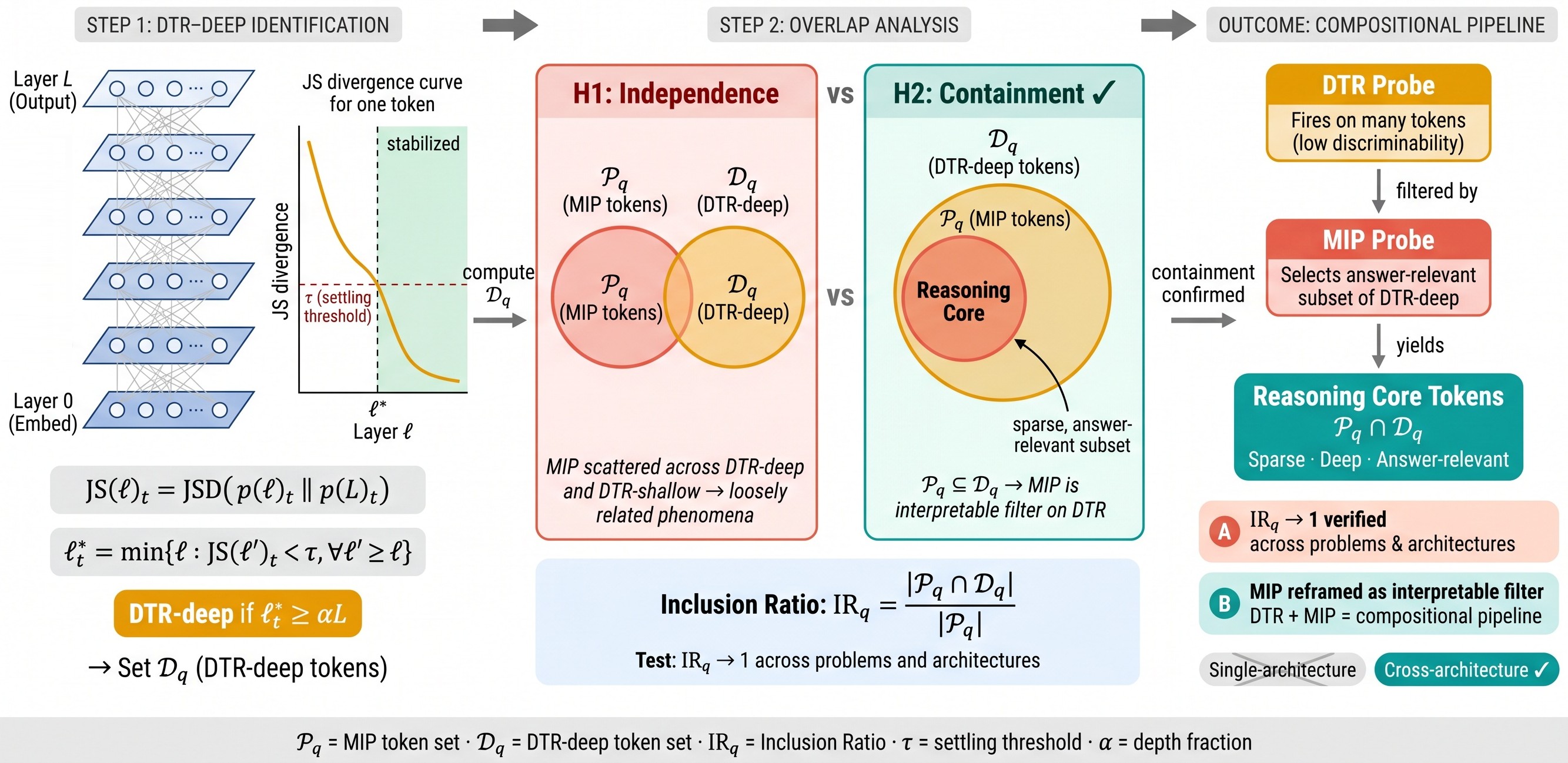}
\vspace{-0.52cm}
\caption{Overview of the proposed overlap analysis.}
\label{Fig:IAR_overlap}
\vspace{-0.36cm}
\end{figure*}

\section{Methodology}
\label{sec:method}

IAR is organized as a three-fold pipeline that jointly analyzes LLM reasoning along three integral axes: \emph{reasoning-relevance}, \emph{computational depth}, and \emph{noise-robustness and reasoning quality}. 

\paragraph{Notation.}
For a problem $q$ with gold answer $y$, let $\mathbf{x} = (x_1, \dots, x_T)$ denote the generated token sequence and $\mathbf{h}^{(\ell)}_t \in \mathbb{R}^{d}$ the hidden state at layer $\ell \in \{1, \dots, L\}$ for token $x_t$. Let $p^{(\ell)}_t(\cdot) = \mathrm{softmax}(W_U \mathbf{h}^{(\ell)}_t)$ be the layer-$\ell$ predictive distribution over the vocabulary obtained by applying the model's unembedding $W_U$. We denote the final-layer distribution by $p^{(L)}_t$.

\subsection{Reasoning-Crucial Token Isolation via
Bandwidth-Calibrated MIP}
\label{sec:step1}

\paragraph{Per-token MI via HSIC.}
For each token $x_t$, we estimate the MI between
its output-layer representation $\mathbf{h}^{(L)}_t$ and the gold answer embedding $\mathbf{e}_y$ via the Hilbert--Schmidt Independence Criterion (HSIC) \citep{gretton2005} under a Gaussian RBF kernel:
\begin{equation}
k_\sigma(\mathbf{u}, \mathbf{v}) =
\exp\!\left( -\tfrac{1}{2\sigma^2}\|\mathbf{u} - \mathbf{v}\|_2^2\right),
\end{equation}
and obtain a per-token MI trace $\mathrm{MI}(x_t) = \widehat{\mathrm{HSIC}}(\mathbf{h}^{(L)}_t, \mathbf{e}_y)$.

\paragraph{Architecture-wise bandwidth calibration.}
A mis-scaled bandwidth $\sigma$ drives all pairwise similarities toward the kernel's saturation regime (\emph{kernel collapse}), destroying the MI signal. We therefore calibrate $\sigma$ per architecture: $\sigma = 50$ for Qwen-7B and Qwen-14B, and a per-problem median heuristic $\sigma_q = \mathrm{median}\{\|\mathbf{h}^{(L)}_i - \mathbf{h}^{(L)}_j\|_2\}_{i<j}$ for Llama-8B. This calibration is validated by verifying that the resulting kernel matrices have non-degenerate spectra across all four domains.

\paragraph{Tukey IQR peak-detection.}
Given the calibrated MI trace, we declare token $x_t$ as a peak if
\begin{equation}
\mathrm{MI}(x_t) \;>\; Q_3 \;+\; 1.5 \cdot \mathrm{IQR},
\end{equation}
where $Q_3$ and $\mathrm{IQR}$ are computed over the per-problem trace. This non-parametric threshold avoids distributional assumptions and adapts automatically to per-problem variance. The output is, for every problem $q$, a sparse set $\mathcal{P}_q \subset \{1, \dots, T\}$ of reasoning-crucial tokens at the output layer.

\subsection{Cross-Layer Trajectory Analysis via MIP--DTR
Token Overlap}
\label{sec:step2}

\paragraph{DTR-deep token identification.}
Apart from reasoning-crucial token isolation in Section \ref{sec:step1}, we compute the layer-wise predictive trajectory for each token by measuring the Jensen--Shannon divergence between the layer-$\ell$ distribution and the final-layer distribution:
\begin{equation}
\mathrm{JS}^{(\ell)}_t \;=\; \mathrm{JSD}( p^{(\ell)}_t \,\|\, p^{(L)}_t ).
\end{equation}
We define the \emph{settling layer} of token $x_t$ as
\begin{equation}
\ell^{\star}_t \;=\; \min \{ \ell \;:\; \mathrm{JS}^{(\ell')}_t < \tau, \;\; \forall \ell' \geq \ell \},
\end{equation}
where $\tau$ is an architecture-specific settling threshold. Token $x_t$ is declared \emph{DTR-deep} if $\ell^{\star}_t \geq \alpha L$ for a depth fraction $\alpha \in (0,1)$, indicating that substantial cross-layer computation was required before the prediction stabilized. Let $\mathcal{D}_q$ denote the resulting set of DTR-deep tokens.

\paragraph{Overlap analysis.}
We then conduct a set-level overlap analysis between
$\mathcal{P}_q$ and $\mathcal{D}_q$ based on
two qualitatively different hypotheses:
\begin{itemize}
    \item \textbf{H1 (Independence):} MIPs are scattered across both DTR-deep and DTR-shallow tokens, where reasoning-relevance and computational depth are loosely related.
    \item \textbf{H2 (Containment):} $\mathcal{P}_q \subseteq \mathcal{D}_q$, in which case MIPs form a \emph{sparse, reasoning-relevant subset of the computationally
    deep tokens}.
\end{itemize}
We quantify containment via the inclusion ratio
\begin{equation}
\mathrm{IR}_q \;=\; |\mathcal{P}_q \cap \mathcal{D}_q|/|\mathcal{P}_q|,
\end{equation}
and test $\mathrm{IR}_q \to 1$ across problems and architectures. A positive containment outcome is non-trivial: DTR-deep alone fires on too many tokens to be a discriminative interpretability signal, so containment establishes MIP as an \emph{interpretable filter on DTR-deep tokens}, selecting which computationally deep tokens are also answer-relevant. This reframes the two probes from competing measurements into a compositional pipeline. The process of this overlap analysis is illustrated in Figure \ref{Fig:IAR_overlap}.

\subsection{Reasoning-Quality Validation via
MIP-Statistics-Driven Jaccard Stability}
\label{sec:step3}

\paragraph{Three-way Jaccard stability.}
To verify that the tokens identified in Section \ref{sec:step1} and \ref{sec:step2} reflect reasoning-quality-guaranteed structure rather than artifacts of a single stochastic sample, we generate three independently seeded reasoning chains for each problem $q$ and compute the corresponding MIP sets $\mathcal{P}_q^{(1)}, \mathcal{P}_q^{(2)}, \mathcal{P}_q^{(3)}$. The three-way Jaccard stability of problem $q$ is then defined as \begin{equation}
J_3(q)\!=\!{\big|\mathcal{P}_q^{(1)} \!\cap\! \mathcal{P}_q^{(2)} \!\cap\! \mathcal{P}_q^{(3)}\big|}/
                   {\big|\mathcal{P}_q^{(1)} \!\cup\! \mathcal{P}_q^{(2)} \!\cup\! \mathcal{P}_q^{(3)}\big|},
\end{equation}
which provides a decoding-variance-robust measure of whether the reasoning-crucial tokens recur across generations queried from same question.

\paragraph{External reasoning-quality taxonomy.}
To evaluate whether MIP statistics genuinely track reasoning quality, we adopt the \emph{Genuine / Lucky / Silent} partition of \citet{sahoo2026} as an external behavioral taxonomy. The partition is defined at the \textbf{problem level}, not the token level. Let $c_q \in \{0,1,2,3\}$ denote the number of correct generations among the three seeded samples of problem $q$. We assign $q$ to a category based on $(c_q, J_3(q))$:
\begin{itemize}
    \item \textbf{Genuine}: $c_q = 3$ and $J_3(q) \geq \tau_J$ (consistently correct \emph{and} stable MIP structure);
    \vspace{-0.2cm}
    \item \textbf{Lucky}: $c_q \geq 1$ and $J_3(q) < \tau_J$
    (sometimes correct but unstable MIP structure);
    \vspace{-0.2cm}
    \item \textbf{Silent}: $c_q = 0$ (never correct, regardless of MIP stability).
\end{itemize}
Our adaptation replaces Sahoo's activation consistency confidence score with this MIP-based stability metric while preserving the original behavioral semantics of the three categories.

\paragraph{Aggregate MIP-statistics discrimination.}
This partition provides a \emph{reasoning-quality signal independent of the MIP statistics under test}. We therefore ask whether three aggregate MIP statistics discriminate Genuine problems from Lucky and Silent ones: peak \emph{count} $|\mathcal{P}_q|$, peak \emph{ratio} $|\mathcal{P}_q|/T$, and peak \emph{intensity} $\bar{\mathrm{MI}}_q = \frac{1}{|\mathcal{P}_q|}\sum_{t \in \mathcal{P}_q}\mathrm{MI}(x_t)$. The question addressed is therefore \emph{not} whether individual tokens contribute to reasoning because tokens are never partitioned into these categories, but whether aggregate MIP-statistic effect sizes hold across reasoning-quality regimes and across architectures. Statistical comparisons use Mann--Whitney $U$ tests with rank-biserial effect sizes, which are non-parametric and robust to the skewed distributions typical of peak-count data.

\section{Experiments}

\subsection{Setup and Implementation}

\paragraph{Models.}
We use six reasoning models: Qwen-7B (28 layers), Qwen-14B (48 layers), Llama-8B (32 layers), Phi-4-mini (32 layers), EXAONE-7.8B (32 layers), and GPT-OSS-20B (24 layers). The Qwen-7B and Qwen-14B share the Qwen-2 architecture, allowing cross-scale evaluation. Llama-8B uses a distinct structure for cross-architecture validation. Phi-4-mini (3.8B parameters) extends the evaluation to the small-model regime. EXAONE-7.8B represents an independently developed architecture and training pipeline outside the Qwen/Llama families. GPT-OSS-20B introduces a sparse Mixture-of-Experts design (32 routed experts with top-4 routing), testing whether our findings transfer from dense to sparse architectures.

\paragraph{Datasets.} 
We evaluate the proposed IAR across four cognitive domains, with 200 problems per domain (800 problems per model): GSM8K \citep{cobbe2021} for mathematics, MBPP \citep{austin2021} for code generation, BBH Boolean Expressions \citep{suzgun2022} for logic, and CommonsenseQA \citep{talmor2019} for commonsense. This setup balances symbolic-gold tasks (math, code) against natural-language-gold tasks (logic, commonsense). For the latter, logic tasks are evaluated using templated boolean sentences, and commonsense tasks utilize ECQA-derived rationale-and-answer text \citep{aggarwal2021}.

\paragraph{Metrics.}

\textbf{WPR}: With-Peaks Rate (\%) of problems yielding at least one Tukey-detected MIP; \textbf{TPP}: Token-Pool Precision $|\mathcal{P}\cap\mathcal{D}|/|\mathcal{P}|$, pooled over the 200 problems per (architecture, domain) pair; \textbf{PPR}: Per-Problem Recall $|\mathcal{P}_q\cap\mathcal{D}_q|/|\mathcal{D}_q|$, averaged over with-peaks problems; \textbf{J3}: three-way Jaccard stability $J_3(q)$ averaged over with-peaks problems; \textbf{NPR}: No-Peak Rate (\%) across all three sampled seeds; \textbf{CCR}: Consistent-Correctness Rate (\%) required three seeds correct; \textbf{MIP-Stats}: rank-biserial effect sizes (Mann--Whitney $U$ test, Genuine vs. Lucky) for peak count / ratio / intensity.

\paragraph{Rationale of Metrics.}
These metrics jointly capture the prevalence, precision, stability, and statistical significance of MIPs within the DTR-deep band. \textbf{WPR}, \textbf{NPR}, and \textbf{CCR} aggregate per-problem outcomes into architecture-level rates, enabling fair comparison across models. \textbf{TPP} and \textbf{PPR} capture the overlap between MIP and DTR-deep token sets, naturally testing whether MIPs fall inside the DTR-deep band, reported in both token-pooled and per-problem-averaged forms to address aggregate concentration and per-instance reliability, respectively. \textbf{J3} extends Jaccard to three seeded generations, thereby testing robustness under stochastic decoding. Finally, \textbf{MIP-Stats} serves as non-parametric choice for unbalanced groups and non-normal peak-count distributions in reasoning quality partitions.

\paragraph{Implementation.} We use greedy decoding to obtain deterministic generations, and use sampling decoding with temperature $T=0.7$ and three random seeds (42, 123, 456) for stochastic generation.  Generations are calibrated per architecture to avoid truncation: 512 tokens for Qwen-7B, and 4096 tokens for Qwen-14B, Llama-8B, Phi-4-mini, EXAONE-Deep-7.8B, and GPT-OSS-20B. In particular, for the code domain, where program traces run longer, the budget is 8192 tokens for all models. Each model uses its native chat template to elicit the full reasoning trace. MI estimation follows the HSIC implementation of \citet{qian2025} with a 512-dimensional subsample of the final-layer hidden state. DTR computation applies the model's final RMSNorm before the unembedding projection ($p^{(\ell)}_t = \mathrm{softmax}(W_U \cdot \mathrm{RMSNorm}(\mathbf{h}^{(\ell)}_t))$), aligning with the logit-lens tradition \citep{belrose2023}; the JSD settling threshold is $\tau = 0.5$ and the depth fraction is $\alpha = 0.85$.

\begin{table*}[tp]
\centering
\footnotesize
\renewcommand{\arraystretch}{0.9}
\setlength{\tabcolsep}{8.38pt}
\begin{tabularx}{\linewidth}{c|c|ccc|cccc}
\toprule
Models & Domains & WPR & TPP & PPR & J3 & NPR & CCR & MIP-Stats \\ \midrule
\multirow{4}{*}{Qwen-7B}  & Math        & 63.00 & 0.77 & 0.01 & 0.23 & 36.00  & 79.50 & -0.83/-0.83/-0.41 \\
                          & Code        & 79.50 & 0.91 & 0.02 & 0.08 & 15.50  & 49.00 & -0.92/-0.91/-0.55 \\
                          & Logic       & 0.00  & --   & --   & --   & 100.00 & 95.00 & {\boldmath$G{=}0$} \\
                          & Commonsense & 1.50  & --   & --   & --   & 99.00  & 54.00 & {\boldmath$G{=}0$} \\ \hline
\multirow{4}{*}{Qwen-14B} & Math        & 35.00 & 0.84 & 0.01 & 0.28 & 53.00  & 94.00 & -0.93/-0.92/-0.96 \\
                          & Code        & 54.00 & 0.87 & 0.02 & 0.12 & 23.50  & 71.00 & -0.88/-0.85/-0.59 \\
                          & Logic       & 100.00& 0.94 & 0.10 & 0.16 & 0.00   & 99.00 & -0.75/-0.98/-1.00 \\
                          & Commonsense & 95.00 & 0.91 & 0.09 & 0.15 & 20.50  & 76.50 & -0.98/-0.98/-0.93 \\ \hline
\multirow{4}{*}{Llama-8B} & Math        & 96.50 & 0.85 & 0.02 & 0.09 & 2.00   & 58.00 & -0.66/-0.67/-0.09 \\
                          & Code        & 87.00 & 0.63 & 0.04 & 0.05 & 12.50  & 36.50 & -0.84/-0.86/-0.40 \\
                          & Logic       & 91.50 & 0.72 & 0.02 & 0.01 & 1.50   & 76.00 & -0.70/-0.66/-0.84 \\
                          & Commonsense & 89.50 & 0.78 & 0.03 & 0.04 & 10.00  & 61.00 & -0.79/-0.77/-0.56 \\ \hline
\multirow{4}{*}{Phi-4-mini} & Math      & 100.00 & 0.98 & 0.01 & 0.20 & 0.00  & 96.00 & {\boldmath$L{=}1$} \\
                          & Code        & 82.50 & 0.96 & 0.01 & 0.05 & 9.00  & 30.50 & -0.72/-0.64/-0.30 \\
                          & Logic       & 59.50 & 0.99 & 0.01 & 0.10 & 23.00 & 97.50 & -0.71/-0.74/-0.85 \\
                          & Commonsense & 83.50 & 0.97 & 0.01 & 0.17 & 6.00  & 41.50 & -0.34/-0.55/-0.63 \\ \hline
\multirow{4}{*}{EXAONE-Deep-7.8B} &Math & 100.00& 0.62 & 0.09 & 0.25 & 0.00  & 95.00 & {\boldmath$L{=}0$} \\
                          & Code        & 100.00& 0.60 & 0.08 & 0.23 & 0.00  & 66.50 & {\boldmath$L{=}0$} \\
                          & Logic       & 100.00& 0.58 & 0.05 & 0.17 & 0.00  & 86.00 & {\boldmath$L{=}1$} \\
                          & Commonsense & 100.00& 0.55 & 0.06 & 0.17 & 0.00  & 57.00 & {\boldmath$L{=}0$} \\ \hline
\multirow{4}{*}{GPT-OSS-20B} & Math     & 64.00 & 0.39 & 0.00 & 0.10 & 13.00 & 95.00 & -0.86/-0.82/-0.78 \\
                          & Code        & 91.00 & 0.44 & 0.01 & 0.03 & 3.50  & 78.00 & -0.68/-0.60/-0.31 \\
                          & Logic       & 51.50 & 0.35 & 0.00 & 0.04 & 31.50 & 97.00 & -0.85/-0.91/-0.93 \\
                          & Commonsense & 56.50 & 0.35 & 0.01 & 0.03 & 21.00 & 77.00 & -0.86/-0.68/-0.87 \\
\bottomrule
\end{tabularx}
\caption{Main results of IAR across six architectures and four domains. Left block: Detectability and Containment (WPR/TPP/PPR) across four domains. Right block: Stability and Quality Discrimination (J3/NPR/CCR/MIP-Stats). Notably, $G$ = Genuine count, $L$ = Lucky count, where Genuine-vs-Lucky is untestable in these cells.}
\label{tab:main_results}
\vspace{-0.2cm}
\end{table*}

\subsection{Main Results and Analyses}
\label{sec:main_results}
We carry out extensive experiments to verify the interpretation capability of the proposed IAR framework. Beyond the three DeepSeek-R1-Distill models, we further instantiate IAR on three additional architectures, i.e., Phi-4-mini, EXAONE-Deep-7.8B, and GPT-OSS-20B (a mixture-of-experts model), yielding six models spanning five distinct architectural families. All results are consolidated in Table~\ref{tab:main_results}.
Rather than presenting numbers cell-by-cell, we organize the empirical findings around four research questions (RQs) that map onto the three axes of IAR (reasoning-relevance, computational depth, noise-robustness) and onto its cross-architecture validation: \textbf{RQ1 (Detectability):} Does bandwidth-calibrated MIP with Tukey IQR peak detection generalize across architectures and domains? \textbf{RQ2 (Containment):} Are MIP-identified tokens a high-precision subset of DTR-deep tokens, i.e.\ does the Containment hypothesis H2 dominate the Independence hypothesis H1? \textbf{RQ3 (Stability):} Does the MIP token set recur across stochastic generations, and how does stability vary by architecture and by domain? \textbf{RQ4 (Quality Discrimination):} Do aggregate MIP statistics (peak count, peak ratio, peak intensity) discriminate \emph{Genuine} reasoning from \emph{Lucky} guesses and \emph{Silent} failures, and is this discrimination architecturally invariant?

\paragraph{RQ1: Detectability of MIP across architectures and domains.}
WPR reveals a pronounced \emph{architecture$\times$domain interaction}, yet detection is broadly achievable: five of six architectures produce peaks in every domain. Llama-8B exhibits a near-uniform WPR (87.00--96.50\%), EXAONE-Deep-7.8B saturates at 100.00\% across all four domains, and Phi-4-mini (59.50--100.00\%) and GPT-OSS-20B (51.50--91.00\%) remain productive throughout, indicating that Tukey-IQR peak detection on the bandwidth-calibrated MI trace is domain-agnostic for these architectures. In contrast, Qwen-7B fails on natural-language-gold tasks (logic: 0.00\%; commonsense: 1.50\%) while remaining productive on symbolic-gold tasks (math: 63.00\%; code: 79.50\%), and Qwen-14B occupies an intermediate regime in which symbolic-gold tasks suppress peak counts (35.00\%, 54.00\%) but natural-language-gold tasks saturate (100.00\%, 95.00\%). Two mechanisms jointly explain this pattern. First, symbolic-gold answers yield concentrated answer embeddings, producing sharply localized MIPs; natural-language-gold answers yield diffuse embeddings that inflate MIP counts in output spaces. Second, the model-wise bandwidth calibration interacts with the spectral properties of the final-layer hidden states: Qwen-7B's smaller embedding dimensionality collapses kernel similarities for natural-language gold answers, explaining its near-zero WPR on logic and commonsense. The expanded evaluation shows this failure mode is architecture-specific rather than intrinsic to the probe: bandwidth-calibrated MIP \emph{detects} reasoning-crucial tokens across families, with sensitivity conditionally coupled to (architecture, domain).

\paragraph{RQ2: Containment of MIP within DTR-deep tokens.}
Across all 22 (model, domain) cells with sufficient peak data, TPP ranges from 0.35 to 0.99 with a pooled mean of $0.73$ and a median of $0.78$, revealing that containment is \emph{graded} rather than binary. Containment is strong-to-near-ceiling for the dense Qwen, Llama, and Phi families (0.63--0.99; Phi-4-mini reaches 0.96--0.99), moderate for EXAONE-Deep-7.8B (0.55--0.62), and weakest for the MoE-based GPT-OSS-20B (0.35--0.44), suggesting that expert routing disperses answer-relevant information outside the DTR-deep band. Nevertheless, H2 dominates H1 in the large majority of cells, and the key asymmetry between high TPP and uniformly low PPR ($\leq 0.10$) persists in \emph{every} architecture: MIP tokens are reliably concentrated in $\mathcal{D}_q$ but constitute only a small minority of it. This is exactly the behavior expected if MIP acts as an \emph{interpretable filter} on DTR, i.e., DTR-deep tokens are necessary but not sufficient for reasoning relevance, and MIP supplies the missing sufficiency condition. Within the Qwen family, the ordering Qwen-14B (TPP $\approx 0.89$) $>$ Qwen-7B ($\approx 0.84$) still indicates that containment strength scales with depth and scale under a fixed architecture, while the cross-family spread identifies containment strength as a family-level property.

\paragraph{RQ3: Token-level stability across stochastic generations.}
J3 has not previously been reported in the output-layer MIP literature. Although TPP is high, J3 is uniformly low across all 22 cells ($\leq 0.28$). The interpretation is \emph{not} that MIP is an unreliable probe, but that the \emph{identity} of MIP tokens is sample-specific. Two pieces of evidence support this re-interpretation. First, NPR is approximately stable across seeds within each (architecture, domain) cell, and EXAONE-Deep-7.8B attains NPR $=0.00\%$ in all four domains, meaning that the binary event of detecting a MIP is reproducible even when the specific peaks are not. Second, the full-domain view reveals a systematic \emph{domain effect}: mathematics yields the highest J3 within every single architecture (e.g., 0.28 for Qwen-14B, 0.25 for EXAONE, 0.23 for Qwen-7B, versus $\leq 0.05$ for GPT-OSS-20B on code, logic, and commonsense), indicating that symbolic reasoning anchors recur across generations far more reliably than natural-language ones. This qualifies our earlier conjecture that peak abundance alone lowers set overlap: EXAONE combines saturated WPR with the highest non-math J3 values, so gold-answer structure, not peak volume, governs identity stability. This decoupling of token-identity stability from peak-statistic stability motivates RQ4: if individual tokens are unstable but aggregate counts and intensities are not, then reasoning quality must be read off the aggregate statistics rather than the tokens themselves.

\paragraph{RQ4: MIP statistics discriminate reasoning-quality regimes.}
MIP-Stats reports rank-biserial effect sizes from Mann--Whitney $U$ tests comparing Genuine problems against Lucky ones, separately for peak count, peak ratio, and peak intensity. Of the 24 cells, 17 admit the test, and \emph{all 51 resulting effect sizes are negative}: Genuine problems exhibit \emph{fewer, sparser, and weaker} MIPs than Lucky ones, a sign-invariant ordering across five architectural families. This inverts the naive expectation that more peaks indicate more reasoning, and is consistent with an emerging \emph{reasoning-economy} view of CoT: problems that the model truly solves require a small number of sharply localized reasoning anchors, whereas Lucky trajectories produce diffuse, high-volume MI activity. Qwen-14B reaches near-ceiling separation ($|r|$ up to $1.00$ on logic intensity, Lucky $n = 3$), and, notably, GPT-OSS-20B remains strongly discriminative ($|r| \in [0.31, 0.93]$) despite its weak containment, showing that quality discrimination does not require tight MIP--DTR containment. The seven untestable cells are themselves informative and arise from two \emph{opposite} mechanisms: Qwen-7B's logic/commonsense cells are degenerate through detection failure ($G{=}0$), whereas EXAONE's cells (and Phi-4-mini's math cell) are degenerate through near-perfect reliability ($L{\leq}1$ with NPR $=0.00\%$ and high CCR), i.e., virtually every solved problem is Genuine. The residual near-null case remains Llama-8B's math intensity ($|r|=0.09$), attributable to its median-heuristic bandwidth, which preserves detection coverage at the cost of intensity-based separability; its peak counts and ratios remain discriminative ($|r| \geq 0.66$), confirming that the qualitative discrimination direction is architecture-invariant.

\paragraph{Cross-architecture synthesis.}
Taken together, the four research questions yield a coherent picture of LLM reasoning that no single probe could establish: (i) MIP detection generalizes to five of six architectures unconditionally, with sensitivity governed by the joint coupling of architecture and domain (RQ1). (ii) Whenever MIP fires, it concentrates inside the DTR-deep band, strongly for dense architectures and in graded form for MoE routing, validating the compositional reframing of MIP as a sparse, reasoning-relevant filter on DTR (RQ2). (iii) Token-level identity is sample-fragile while peak-event existence is sample-stable, with symbolic domains anchoring the most recurrent MIPs (RQ3). (iv) Aggregate MIP statistics discriminate Genuine from Lucky/Silent reasoning with negative effect sizes in \emph{all} testable cells (RQ4). The two robust phenomena (i.e., MIP--DTR containment and the Genuine $<$ Lucky ordering of MIP statistics) replicate across five architectural families, including a mixture-of-experts model, providing the first cross-architecture evidence of a transferable reasoning pattern in LLMs.

\subsection{Ablation Study}
\label{sec:ablation}
We conduct two types of ablation studies. The first examines the sensitivity of the MIP detection to the HSIC kernel bandwidth $\sigma$ (Table \ref{tab:sigma_ablation}), while the second probes the stability of reasoning-quality partition with respect to the threshold $\tau_J$ (Table \ref{tab:rq4_ablation}). 

\subsubsection{Sensitivity to HSIC Kernel Bandwidth $\sigma$}
\label{sec:sigma_ablation}

\begin{table}[t]
\centering
\footnotesize
\renewcommand{\arraystretch}{1.0} 
\setlength{\tabcolsep}{5pt}
\begin{tabularx}{\linewidth}{cccccc}
\toprule
\textbf{Domain} & \textbf{10} & \textbf{25} & \textbf{50} & \textbf{100} & \textbf{200} \\
\midrule
Math         & 52.00 & 52.00 & 62.50 & 82.50  & 94.00  \\
Code         & 89.50 & 91.00 & 94.00 & 100.00 & 100.00 \\
Logic        & 41.50 & 0.00  & 0.00  & 22.00  & 60.50  \\
Commonsense  & 46.50 & 9.50  & 4.00  & 22.00  & 88.00  \\
\bottomrule
\end{tabularx}
\caption{Sensitivity of \textbf{WPR} to the HSIC kernel bandwidth $\sigma$ on Qwen-7B. }
\label{tab:sigma_ablation}
\vspace{-0.3cm}
\end{table}

\paragraph{Symbolic-gold domains are $\sigma$-robust.}
For both math and code, WPR is non-decreasing in $\sigma$, and detection never collapses across the tested range. The baseline $\sigma = 50$ falls within a stable interior regime for both domains, indicating that the main-paper findings are not contingent on a fine-tuned bandwidth choice. We do, however, observe a systematic shift in the lexical composition of detected peaks: at narrow bandwidths ($\sigma \leq 25$), peaks predominantly fire on punctuation and whitespace tokens, whereas at wider bandwidths ($\sigma \geq 100$), they concentrate on semantically meaningful reasoning markers (e.g., \texttt{So}, \texttt{Let}, \texttt{That}). This trend is consistent with the kernel acting as a smoothing operator that increases the relative salience of semantically informative tokens above the IQR threshold.

\paragraph{Natural-language-gold domains are $\sigma$-non-monotonic.}
In contrast, the logic and commonsense domains exhibit a pronounced trough centered near $\sigma = 50$. Substantial peak coverage is observed at both ends of the bandwidth range ($41.5\%$ and $46.5\%$ at $\sigma = 10$; $60.5\%$ and $88.0\%$ at $\sigma = 200$), with near-complete suppression at intermediate values. At wider bandwidths, the recovered peaks fire predominantly on reasoning markers (\texttt{The}, \texttt{So}, \texttt{I}), with semantic content comparable to those detected in the symbolic-gold domains. 

\subsubsection{Sensitivity to the Stability Threshold $\tau_J$}
\label{sec:rq4_threshold_ablation}

\paragraph{Monotonic attenuation establishes a graded effect.}
Across all architectures, the Genuine-vs-Lucky effect size $|r|$ decreases monotonically as $\tau_J$ tightens. Averaged across models, the peak-count effect attenuates from $|r| = 0.81$ at baseline to $0.68$ and $0.51$ under the two stricter thresholds (relative reductions of $16\%$ and $37\%$). The directional effect persists at every cell, showing that Genuine--Lucky partition is a graded property of the underlying MI dynamics rather than a thresholding artifact.

\begin{table}[t]
\centering
\footnotesize
\renewcommand{\arraystretch}{0.9} 
\setlength{\tabcolsep}{1pt}
\begin{tabularx}{\linewidth}{cccccccc}
\toprule
\textbf{Setting} & \textbf{Models} & \textbf{\makecell[c]{Gen\\uine}} & \textbf{\makecell[c]{Luc\\ky}} & \textbf{\makecell[c]{Sil\\ent}} & \textbf{\makecell[c]{Peak\\Count}} & \textbf{\makecell[c]{Peak\\Ratio}} & \textbf{\makecell[c]{Peak\\Intensity}} \\
\midrule
\multirow{3}{*}{\makecell[c]{Baseline \\($\tau_J > 0$)}}
 & Qwen-7B  & 58 & 101 & 41 & $-0.83$ & $-0.83$ & $-0.41$ \\
 & Qwen-14B & 68 & 120 & 12 & $-0.93$ & $-0.92$ & $-0.96$ \\
 & Llama-8B & 68 & 48  & 84 & $-0.66$ & $-0.67$ & $-0.09$ \\
\midrule
\multirow{3}{*}{\makecell[c]{Strict \\($\tau_J \geq 0.1$)}}
 & Qwen-7B  & 53 & 106 & 41 & $-0.76$ & $-0.77$ & $-0.43$ \\
 & Qwen-14B & 56 & 132 & 12 & $-0.85$ & $-0.85$ & $-0.86$ \\
 & Llama-8B & 50 & 66  & 84 & $-0.42$ & $-0.46$ & $-0.06$ \\
\midrule
\multirow{3}{*}{\makecell[c]{Stricter \\($\tau_J \geq 0.2$)}}
 & Qwen-7B  & 35 & 124 & 41 & $-0.49$ & $-0.50$ & $-0.53$ \\
 & Qwen-14B & 39 & 149 & 12 & $-0.66$ & $-0.65$ & $-0.75$ \\
 & Llama-8B & 20 & 96  & 84 & $-0.39$  & $-0.37$  & $-0.08$ \\
\bottomrule
\end{tabularx}
\caption{Sensitivity of the reasoning quality partition and Genuine-vs-Lucky rank-biserial effect size to the stability threshold $\tau_J$ on the math domain. Notably, owing to space limitation, this table just illustrates the results of math domain and other three domains' results are detailed at Appendix \ref{sec:abl_full_x}.}
\label{tab:rq4_ablation}
\vspace{-0.2cm}
\end{table}

\paragraph{Reclassification asymmetry separates effect-size decay from power loss.}
Threshold tightening induces strictly unidirectional Genuine$\to$Lucky migration, but at model-specific rates. By defining the reclassification rate $\rho\!=\!(G_{\text{Baseline}}-G_{\text{Stricter}}) / G_{\text{Baseline}}$, we obtain $\rho = 0.40$, $0.43$, and $0.71$ for Qwen-7B, Qwen-14B, and Llama-8B. Llama-8B's high $\rho$ shrinks its Genuine group to $20$, at which point its non-trivial effect ($0.39$) fails the Bonferroni correction not because the effect has vanished but because power has collapsed. This distinction matters for replication: threshold-induced significance loss can reflect sample attrition rather than failure of the discriminator, and reporting effect sizes across thresholds is therefore essential.

\section{Conclusion}

We presented IAR to probe how LLMs reason along reasoning-relevance (bandwidth-calibrated MIPs), computation depth (MIP trajectories), and noise-robustness (reasoning quality partition). Instantiating IAR on six models across four domains yields cross-architecture findings: (i) MIPs are semantically meaningful across model-domain pairs under $\sigma$ calibration; (ii) MIPs generally lie inside DTR-deep tokens, establishing MIP as an interpretable filter on the DTR signal; (iii) and the reasoning quality discrimination via MIP statistics is portable across architectures. 

\section{Limitations}

IAR identifies MIPs where answer-relevant alignment occurs along the generation trajectory but does not specify which internal features drive that alignment. Because feature-level analyses identify what concepts the model truly represents and learns, integrating our position-level analysis with feature-level mechanistic decompositions is synergistic. The two analyses operate at different granularities (feature-level versus token-position-level) and could be combined in future work to localize specific features to MIP positions, further disclosing the internal reasoning mechanism in LLMs.

\section{Acknowledgements} 
This work was partly supported by the National Natural Science Foundation of China (Grant No. 62576191), the Shenzhen Science and Technology Program (Grant No. ZDCY20250901103533010), and China Postdoctoral Science Foundation Program (Grant No. 2026M791759).

\appendix

\section{Supplementary Materials for Main Results and Analysis}
\label{app:stats}

This appendix reports the full statistical details for the cross-architecture comparisons summarized in Section~\ref{sec:main_results}. The main paper reports headline numbers (effect sizes, with-peaks rates, partition counts) but compresses the full test family into summary annotations such as ``survives Bonferroni'' or significance asterisks. This appendix unpacks those annotations so that every claim in the main paper can be traced to a specific statistical test with its raw $U$ or $\chi^2$ statistic, uncorrected $p$-value, Bonferroni-corrected verdict, and (for continuous-statistic tests) effect size with 95\% bootstrap confidence interval. All $p$-values reported below are uncorrected; the Bonferroni threshold $\alpha$ applied to each test family is stated at the start of the corresponding subsection. We organize this appendix by research question, mirroring the structure of Section~\ref{sec:main_results}.

\subsection{RQ1: Cross-Architecture Detection}

\paragraph{Detection rates.} The RQ1 block of Table~\ref{tab:main_results} reports the percentage of problems with at least one IQR-detected MI peak across all 24 (model $\times$ domain) cells, of which 22 yield sufficient peaks for analysis. Under per-architecture $\sigma$ calibration, peak detection is viable in every cell of the Llama-8B column (87.0\%--96.5\%), including the natural-language-gold domains (logic, commonsense) where the fixed $\sigma=50$ bandwidth produced essentially no IQR peaks at Qwen-7B. The Qwen-2 family shows a different per-domain pattern: Qwen-7B detects    
peaks in symbolic-gold domains (math 63.0\%, code 79.5\%) but not in NL-gold (logic 0.0\%, commonsense 1.5\%), while Qwen-14B detects peaks in NL-gold at high rates (logic 100\%, commonsense 95.0\%) but at lower rates in symbolic-gold (math 35.0\%, code 54.0\%). The response-period families extend this picture: EXAONE-Deep-7.8B saturates at 100\% in all four domains, Phi-4-mini remains productive throughout (59.5\%--100\%), and GPT-OSS-20B detects peaks in 51.5\%--91.0\% of problems per domain.

\paragraph{Comparability caveat.} These percentages are not directly comparable across architectures because each row's three numbers are produced under a different kernel bandwidth. Per-architecture $\sigma$ calibration places each architecture's MI distribution on its own absolute scale, so the IQR threshold operates on different numerical regimes across columns. The qualitatively comparable claim is detection viability: under per-architecture $\sigma$, the IQR peak detection rule registers semantically meaningful peaks in every (architecture, domain) cell where the gold representation is well-formed.

\paragraph{Peak vocabulary.} Across all three architectures, peak tokens cluster into reasoning-relevant semantic classes: structural commitment markers (\texttt{First}, \texttt{Total}, \texttt{Next}, \texttt{Finally}), reasoning pivots (\texttt{So}, \texttt{Wait}, \texttt{Okay}, \texttt{Hmm}, \texttt{Let}), and answer-relevant tokens (boolean \texttt{True}/\texttt{False} in logic; the multiple-choice marker \texttt{Option} in commonsense). The specific tokens differ across architectures: Llama-8B math is dominated by structural markers (\texttt{Total}, \texttt{First}, \texttt{Next}, \texttt{Finally}, \texttt{Calculate}); Qwen-14B math is dominated by reasoning pivots (\texttt{So}, \texttt{Let}, \texttt{Okay}, \texttt{That}); but the semantic class is consistent across architectures.

\paragraph{Distribution shape.} Per-token MI distributions exhibit architecture-specific shapes. At Qwen-7B, symbolic-gold domains produce sparse-spike unimodal distributions and NL-gold domains produce pooled-bimodal distributions. At Qwen-14B, NL-gold domains transition to per-problem unimodal long-tailed distributions in which 14--15\% of tokens already lie above MI 0.9, while pooled-domain bimodality persists from between-problem clustering. At Llama-8B under the median-heuristic $\sigma$, all four domains produce distributions concentrated above the unit-MI region (40--53\% of tokens above MI 0.9), reflecting the heuristic's design rather than denser sparse-spike behavior.

\paragraph{Chain-opener token identity.} The token occupying position zero of the reasoning trace is highly stable across problems within each architecture: \texttt{First} at Qwen-7B (192/200 problems), \texttt{Okay} at Qwen-14B (200/200), and \texttt{Ċ} (the tokenizer's newline) at Llama-8B (200/200). The pattern replicates on the response-period families: \texttt{<think} at Phi-4-mini and \texttt{Okay} at EXAONE-Deep-7.8B occupy position zero with the same reliability, while GPT-OSS-20B reserves position zero for harmony control tokens, so the anchor analysis does not apply there. The lexical instantiation differs across architectures, but each token sits in the reasoning-pivot or structural-marker class identified above, indicating that the reasoning trace consistently begins with an architecture-specific reasoning anchor. The peak-detection rate at this position varies (27\%--58\% under the IQR rule), so the cross-architecture observation is one of token-identity stability rather than guaranteed peak firing.

\paragraph{Summary.} The IQR peak detection rule generalizes across all three architectures and all four cognitive domains under per-architecture $\sigma$ calibration, with one exception: Qwen-7B on natural-language-gold domains (logic, commonsense), where the fixed $\sigma=50$ bandwidth produces near-zero peak detection. In every other (architecture, domain) cell, the detection method is shape-relative rather than scale-dependent: it fires on right-tail outliers of each problem's MI distribution regardless of the absolute scale at which the distribution sits. Cross-architecture differences in absolute peak rate, peak ratio, and peak intensity are confounded by the $\sigma$ choice and are not interpreted; the interpretable cross-architecture statements concern detection viability, peak vocabulary semantics, and distribution-shape patterns, all of which generalize across the cells where detection is viable.

\subsection{RQ2: Token-Pool Precision Pairwise Tests}
\label{app:rq2_stats}

RQ2 in the main paper claims that the MIP-within-DTR-deep containment relationship is preserved across architectures, with the strongest result being statistical indistinguishability between Qwen-14B (0.84) and Llama-8B (0.85) precision on math ($p = 0.95$). For the three original architectures, this claim rests on a family of 12 pairwise two-proportion $z$-tests across the four domains, plus four three-architecture chi-square contingency tests for overall significance, reported below. The response-period families (Phi-4-mini, EXAONE-Deep-7.8B, GPT-OSS-20B) are instead quantified against a chance baseline via the lift-over-coverage analysis reported in the main text, which subsumes pairwise precision comparisons for those models; their  
containment outcomes, including the GPT-OSS failure and the Phi-4-mini saturation bound, are discussed there. The Bonferroni threshold for the pairwise family is $\alpha = 0.05/12 = 0.0042$.

\paragraph{Overall significance per domain.}
Before testing specific pairwise differences, we first verify that the three architectures are not all identical in any domain. Table~\ref{tab:rq2_chi2} reports three-architecture chi-square contingency tests on the (overlap-with-DTR-deep, no-overlap) classification of pooled peak tokens. The chi-square test treats the three architectures as the row variable and the binary overlap classification as the column variable, asking whether overlap rates differ across the three rows. Every domain rejects the null hypothesis at $p < 10^{-4}$ or stronger. The chi-square magnitude scales with the cross-architecture precision spread: math has the tightest precision range (0.773 to 0.845, span 0.072) and the smallest chi-square ($22.2$); logic has the widest range (0.937 vs 0.724, span 0.213) and the largest chi-square ($1{,}655.7$). The overall result establishes only that the three precision values are not all equal; the pairwise tests in Table~\ref{tab:rq2_pairwise} localize where the differences sit.

\begin{table}[!htbp]
\centering
\small
\setlength{\tabcolsep}{9pt}
\begin{tabularx}{\linewidth}{lccc}
\toprule
\textbf{Domain} & \textbf{$\chi^2$} & \textbf{$p$} & \textbf{Verdict} \\
\midrule
Math         & 22.22    & $<10^{-4}$   & differ \\
Code         & 550.70    & $<10^{-120}$ & differ \\
Logic        & 1655.70   & $<10^{-50}$  & differ \\
Commonsense  & 455.90    & $<10^{-99}$  & differ \\
\bottomrule
\end{tabularx}
\caption{RQ2 three-architecture chi-square contingency tests on token-pool precision per domain ($\mathrm{dof}=2$ each). All four domains reject the null hypothesis that precision is equal across the three architectures.}
\label{tab:rq2_chi2}
\end{table}

\paragraph{Pairwise architecture comparisons.}
With the overall significance established, Table~\ref{tab:rq2_pairwise} reports the 12 pairwise two-proportion $z$-tests that localize the within-domain differences. The central finding is the math row's third entry: Qwen-14B vs Llama-8B yields $z = -0.07$ and $p = 0.95$, the only test in the entire 12-test family that fails to reach significance. This negative result is the cleanest cross-architecture transfer signal in the paper, because it shows that two architectures with different lineages and different $\sigma$ calibrations produce essentially identical precision (0.84 vs 0.85) on the same dataset. Every other pairwise test survives Bonferroni correction by a wide margin, with $z$-statistics ranging from $\pm 3.85$ to over $\pm 18$. The within-Qwen comparisons confirm that precision rises with scale on math (0.77 to 0.84) but falls with scale on code (0.91 to 0.87), refining the simple ``larger is better'' interpretation. The Qwen-vs-Llama comparisons confirm that Llama-8B precision drops dramatically on code (0.91/0.87 to 0.63) and moderately on logic (0.94 to 0.72) and commonsense (0.91 to 0.78), with all four cross-family differences surviving Bonferroni by orders of magnitude.

\paragraph{Containment relationship.} Across all three architectures, MI peaks are with high precision a subset of DTR-deep tokens. The RQ2 block of Table~\ref{tab:main_results} reports token-pool precision (the fraction of pooled MI peak tokens across all 200 problems per cell that are also DTR-deep), which we adopt as the primary cross-architecture measure because it weights tokens equally rather than problems and is independent of per-problem averaging conventions. Qwen-7B logic and commonsense produce too few IQR peaks (0/200 and 4/200) for a meaningful comparison and are excluded.

\paragraph{Math containment is architecture-robust.} Token-pool precision on math is statistically indistinguishable between Qwen-14B (0.844) and Llama-8B (0.845, two-proportion $z$-test $p=0.95$). This is the strongest cross-architecture finding in the precision matrix: on math, the containment relationship transfers cleanly from a Qwen-2 model to a Llama model with a different architectural lineage and a different $\sigma$ calibration, and produces precision values that are essentially identical. The 7B to 14B jump within the Qwen family is also significant after Bonferroni correction at $\alpha=0.0042$ (12 pairwise comparisons), with the larger Qwen scale producing higher precision (0.773 to 0.844). Three-architecture chi-square tests confirm overall significance in every domain ($\chi^2 \geq 22.2$, $p < 10^{-4}$), and 7 of 8 within-domain pairwise $z$-tests survive Bonferroni; the only non-significant pair is the Qwen-14B vs Llama-8B math comparison.

\paragraph{Code precision drops on Llama-8B.} Token-pool precision on code drops substantially at Llama-8B (0.91 to 0.87 to 0.63), with both Qwen-vs-Llama $z$-tests at $p < 10^{-50}$ surviving Bonferroni. Two non-exclusive interpretations are consistent with the data: (i) Llama-8B code peak vocabulary shifts from punctuation and operator tokens (\texttt{,}, \texttt{.}, \texttt{=}, \texttt{==} at Qwen-7B) to reasoning pivots (\texttt{Wait}, \texttt{So}, \texttt{Okay}, \texttt{Let}), and reasoning pivots may settle at mid-depth layers more often than syntactic operators do; (ii) the median-heuristic $\sigma$ on Llama-8B identifies peaks on a larger fraction of problems on code (87.0\% vs 54.0\% with-peaks rate at Qwen-14B), inflating the precision denominator with tokens that fall outside the deep band. The data we have do not separate these interpretations; we flag the code precision drop as a candidate non-generalization.

\paragraph{Natural-language-gold containment.} For logic and commonsense, the cross-architecture comparison is between Qwen-14B and Llama-8B only, because Qwen-7B produces too few IQR peaks under the calibrated NL-gold representation. The containment relationship holds at both architectures but with attenuated precision at Llama-8B (logic: 0.94 to 0.72; commonsense: 0.91 to 0.78). Both differences survive Bonferroni. Llama-8B precision in NL-gold sits well above the chance level, given that DTR-deep covers 47--67\% of tokens, so the relationship generalizes in form even where it does not match Qwen-14B in magnitude.

\paragraph{Recall is uniformly low.} Per-problem recall (the fraction of DTR-deep tokens that are also MI peaks) is below 0.10 in every (architecture, domain) cell. DTR-deep covers 46--76\% of all generated tokens at all three architectures, while MI peaks cover only 1--3\%, so the containment relationship is one-directional rather than mutual. This precision-coverage asymmetry generalizes across architectures: MI peaks identify a sparse, answer-relevant subset of the broad set of computationally demanding tokens that DTR identifies.

\paragraph{Vocabulary disjointness.} The two metrics fire on overlapping token \emph{positions} but on disjoint token \emph{vocabularies}. MI peak vocabulary is dominated by reasoning markers (\texttt{So}, \texttt{Wait}, \texttt{Okay}) and structural commitment words (\texttt{First}, \texttt{Total}, \texttt{Calculate}), while DTR-deep vocabulary is dominated by spacing, punctuation, and structural function words (e.g., \texttt{,}, \texttt{.}, \texttt{the}). The top-20 intersection is below 50\% in every (architecture, domain) cell. This characterization, established at Qwen-7B in prior work, generalizes to all four domains at Qwen-14B and Llama-8B, with logic at Llama-8B as a partial exception where boolean answer tokens enter both vocabularies. The disjointness is consistent with the broader observation that feed-forward layers tend to promote function-word distributions over content-word distributions \citep{geva2021}, while MI peaks fire at content-bearing semantic pivots.

\paragraph{Summary.} The MI-peak-within-DTR-deep containment relationship is preserved across all three architectures, with math precision statistically indistinguishable between Qwen-14B and Llama-8B at $p=0.95$. This is the central cross-architecture finding of the paper: the answer-relevance signal captured by MI peaks is contained within the computational-depth signal captured by DTR, and that containment is a property of how reasoning models commit to answer-relevant tokens at deep layers rather than an artifact of the Qwen-2 architecture. Cross-architecture precision differences in the other three domains (code, logic, commonsense) refine this picture: the containment relationship generalizes in form, with quantitative attenuation on Llama-8B that warrants further investigation.

\begin{table*}[!htbp]
\centering
\small
\setlength{\tabcolsep}{13pt}
\begin{tabularx}{\linewidth}{ccccccc}
\toprule
\textbf{Domain} & \textbf{Pair} & \textbf{$p_a$} & \textbf{$p_b$} & \textbf{$z$} & \textbf{$p$} & \textbf{Bonferroni} \\
\midrule
\multirow{3}{*}{Math}
 & Qwen-7B vs Qwen-14B  & 0.77 & 0.84 & $-3.94$  & $8.20 \times 10^{-5}$  & survives \\
 & Qwen-7B vs Llama-8B  & 0.77 & 0.85 & $-4.26$  & $2.10 \times 10^{-5}$  & survives \\
 & Qwen-14B vs Llama-8B & 0.84 & 0.85 & $-0.07$  & $0.95$                & ns \\
\midrule
\multirow{3}{*}{Code}
 & Qwen-7B vs Qwen-14B  & 0.91 & 0.87 & $+3.85$  & $1.20 \times 10^{-4}$  & survives \\
 & Qwen-7B vs Llama-8B  & 0.91 & 0.63 & $+18.99$ & $<10^{-50}$           & survives \\
 & Qwen-14B vs Llama-8B & 0.87 & 0.63 & $+17.63$ & $<10^{-50}$           & survives \\
\midrule
Logic         & Qwen-14B vs Llama-8B & 0.94 & 0.72 & $+21.94$ & $<10^{-50}$ & survives \\
Commonsense   & Qwen-14B vs Llama-8B & 0.91 & 0.78 & $+14.92$ & $<10^{-40}$ & survives \\
\bottomrule
\end{tabularx}
\caption{RQ2 pairwise two-proportion $z$-tests on token-pool precision, with Bonferroni $\alpha = 4.2 \times 10^{-3}$ matching the 12-cell precision factorial (3 architecture pairs $\times$ 4 domains). Logic and commonsense compare Qwen-14B vs Llama-8B only because Qwen-7B has too few IQR peaks in those domains (0/200 and 4/200, respectively) for a meaningful precision estimate; 8 of the 12 theoretical pairs are computed.}
\label{tab:rq2_pairwise}
\end{table*}

\subsection{RQ3: Sampling Stability Test Family}
\label{app:rq3_stats}

RQ3 in the main paper covers four distinct phenomena that we group into a single 14-test family at Bonferroni $\alpha = 0.05/14 \approx 3.57 \times 10^{-3}$. The phenomena are: (i) the architecture-dependent no-peak rate, tested with 1 three-architecture chi-square plus 3 pairwise $z$-tests; (ii) the correct-all-3 rate, tested with the same chi-square plus pairwise structure; (iii) the three-way Jaccard stability magnitude, tested with 3 pairwise Mann--Whitney $U$ tests on the with-peaks subsets; and (iv) the within-architecture relationship between Jaccard stability and correct-all-3 status, tested with 3 within-architecture Mann--Whitney tests. The 14-test count summing to $(1+3) + (1+3) + 3 + 3$ motivates the Bonferroni threshold.

\paragraph{Scope.} The detailed test family in this subsection is reported for the math domain on the three original architectures, where the IQR rule is best calibrated and gold extraction is exact. The full cross-domain extension of RQ3 to code, logic, and commonsense across all six models, enabled by execution-based correctness for code (standard MBPP practice) and raw-label final-answer extraction for the NL-gold domains, is summarized in the main text and in Table~\ref{tab:rq4_ablation_full}.

\paragraph{Headline numbers.} The RQ3 block of Table~\ref{tab:main_results} reports the headline statistics across architectures. Two patterns are immediately visible: the with-peaks rate (1 minus the no-peak rate, i.e., the fraction of problems with at least one peak across the three seeds) and the with-peaks subset Jaccard move in opposite directions across architectures. Llama-8B sits at the high-density / low-stability extreme (98\% with peaks, mean $J_3 = 0.09$); Qwen-14B sits at the low-density / high-stability extreme (47\% with peaks, mean $J_3 = 0.28$); Qwen-7B sits between on both axes. The use of multiple sampling seeds to assess reasoning robustness is methodologically related to self-consistency decoding \citep{wang2023selfconsistency}, although our analysis targets the recurrence of internal peak signatures rather than agreement among final answers.

\paragraph{No-peak rate across architectures.}
The fraction of math problems for which IQR thresholding registers no peaks under any of three sampling seeds varies dramatically across the three architectures (Qwen-7B 36.0\% / Qwen-14B 53.0\% / Llama-8B 2.0\%). Panel (a) of Table~\ref{tab:rq3_combined} reports the chi-square overall test and the three pairwise $z$-tests for these proportions. The three-architecture chi-square is the strongest single test in the entire RQ3 family ($\chi^2 = 127.6$, $p < 10^{-28}$). All three pairwise comparisons survive Bonferroni. Llama-8B's 2.0\% rate is the dominant signal in the matrix, sitting more than an order of magnitude below either Qwen scale. The Qwen-7B-to-Qwen-14B trend (36\% to 53\%) is also Bonferroni-significant in this test family, but the main-paper interpretation treats this as a within-Qwen pattern rather than a universal scaling signature, given that the rate reverses dramatically at Llama-8B rather than continuing to rise.

\begin{table}[!t]
\centering
\small
\setlength{\tabcolsep}{3pt}
\begin{tabularx}{\linewidth}{lccc}
\toprule
\textbf{Test} & \textbf{Statistic} & \textbf{$p$} & \textbf{Bonf.} \\
\midrule
\multicolumn{4}{l}{\textit{(a) No-peak rate}} \\
3-arch $\chi^2$       & $127.60$    & $<10^{-28}$           & differ \\
Qwen-7B vs Qwen-14B   & $z=-3.42$  & $6.20 \times 10^{-4}$  & surv. \\
Qwen-7B vs Llama-8B   & $z=+8.67$  & $<10^{-15}$           & surv. \\
Qwen-14B vs Llama-8B  & $z=+11.42$ & $<10^{-25}$           & surv. \\
\midrule
\multicolumn{4}{l}{\textit{(b) Correct-all-3 rate}} \\
3-arch $\chi^2$       & $74.50$     & $<10^{-16}$           & differ \\
Qwen-7B vs Qwen-14B   & $z=-4.28$  & $1.90 \times 10^{-5}$  & surv. \\
Qwen-7B vs Llama-8B   & $z=+4.64$  & $3.50 \times 10^{-6}$  & surv. \\
Qwen-14B vs Llama-8B  & $z=+8.43$  & $<10^{-15}$           & surv. \\
\midrule
\multicolumn{4}{l}{\textit{(c) Three-way Jaccard ($U$ test)}} \\
Qwen-7B vs Qwen-14B   & $U=4{,}97$  & $0.02$              & dir. \\
Qwen-7B vs Llama-8B   & $U=14{,}34$ & $0.02$              & dir. \\
Qwen-14B vs Llama-8B  & $U=12{,}59$ & $2.10 \times 10^{-7}$ & surv. \\
\bottomrule
\end{tabularx}
\caption{RQ3 cross-architecture tests ($n=200$ math problems). Bonferroni $\alpha = 0.05/14 \approx 3.57 \times 10^{-3}$. ``surv.'' = survives Bonferroni; ``dir.'' = directional (uncorrected $p<0.05$, fails Bonferroni). Panel (c) reports rank-biserial $r = +0.17, -0.14, -0.37$ respectively.}
\label{tab:rq3_combined}
\end{table}

\paragraph{Correct-all-3 rate across architectures.}
The fraction of problems for which all three seeded generations produce correct answers also varies substantially across the three architectures (Qwen-7B 79.5\% / Qwen-14B 94.0\% / Llama-8B 58.0\%). Panel (b) of Table~\ref{tab:rq3_combined} reports the corresponding chi-square overall test and three pairwise comparisons. All three pairwise differences survive Bonferroni. The correct-all-3 rate is mechanically related to the partition counts reported in RQ4: Qwen-14B's high correct-all-3 rate drives its Lucky-dominated partition (60\% Lucky), while Llama-8B's low correct-all-3 rate drives its Silent-dominated partition (42\% Silent). The cross-architecture correctness variation is itself a confound for cross-architecture stability comparisons, because the Lucky and Silent groups are defined in part by correctness. The main paper handles this confound by restricting cross-architecture stability claims to the with-peaks subset rather than the full sample.

\paragraph{Jaccard stability magnitude.}
The three-way Jaccard stability $J_3$ measures whether the same peak tokens recur across three independently seeded generations. Panel (c) of Table~\ref{tab:rq3_combined} reports the three pairwise Mann--Whitney $U$ tests on $J_3$ distributions, restricted to the with-peaks subsets of each architecture. These are the weakest pairwise findings in the RQ3 family. Only the Qwen-14B vs Llama-8B comparison survives Bonferroni at $\alpha = 3.57 \times 10^{-3}$ ($U = 12{,}585$, $p = 2.1 \times 10^{-7}$, $r = -0.37$), confirming that Qwen-14B has substantially higher stability (mean $J_3 = 0.28$) than Llama-8B (mean $J_3 = 0.09$) on the with-peaks subsets. The Qwen-7B vs Qwen-14B and Qwen-7B vs Llama-8B comparisons reach uncorrected significance only ($p = 0.024$ and $p = 0.022$ respectively), which we flag as \emph{directional}: the effect sign is consistent with the broader cross-architecture pattern, but the comparison does not reach the corrected threshold within this 14-test family. The main-paper claim that ``stability rises at the larger Qwen scale'' is therefore reframed as a directional observation under multi-comparison correction, rather than as an independently established cross-scale signal.

\paragraph{Density-stability tradeoff.} A descriptive pattern emerges across the three architectures: architectures with a higher with-peaks rate produce a lower mean Jaccard, and vice versa. With $n=3$ architectures, this is descriptive rather than an established tradeoff, but it has methodological implications: cross-architecture comparisons of stability should be reported on the with-peaks subset rather than on the full 200 problems, since imputing $J_3 = 0$ for no-peak problems would conflate the no-peak phenomenon with the within-with-peaks-stability phenomenon.

\paragraph{Within-architecture correctness $\times$ stability.}
A natural follow-up question to the cross-architecture stability comparisons above is whether, within each architecture, the three-way Jaccard stability discriminates correct-all-3 problems from not-all-correct problems. Table~\ref{tab:rq3_corrstab} reports the three within-architecture Mann--Whitney tests. None reaches Bonferroni significance at $\alpha = 3.57 \times 10^{-3}$, and none reaches even uncorrected significance ($p$-values 0.069 to 0.31). The Llama-8B test is the most statistically powerful of the three ($n_c = 115$, $n_{nc} = 81$) and is the closest to threshold ($p = 0.069$), but the directional signal is small ($r = -0.15$). This is a meaningful null result: stability alone is not a robust within-architecture discriminator of correctness at any of the three architectures, even though stability does discriminate the partition-defined Genuine and Lucky groups (by construction). This null result motivates RQ4's combination of stability with peak-statistic aggregations rather than relying on stability alone as a single-axis correctness signal.

\begin{table}[!htbp]
\centering
\small
\setlength{\tabcolsep}{7pt}
\begin{tabular}{lcccc}
\toprule
\textbf{Architecture} & \textbf{$n_c / n_{nc}$} & \textbf{$U$} & \textbf{$p$} & \textbf{$r$} \\
\midrule
Qwen-7B   & $101 / 27$ & $1{,}527.50$ & $0.31$  & $-0.12$\\
Qwen-14B  & $89 / 5$   & $298.00$     & $0.20$  & $-0.34$ \\
Llama-8B  & $115 / 81$ & $5{,}336.00$ & $0.07$& $-0.15$ \\
\bottomrule
\end{tabular}
\caption{RQ3 within-architecture correctness $\times$ stability tests. $n_c$ is the number of correct-all-3 problems and $n_{nc}$ is the number of not-all-correct problems. None of the three tests reaches Bonferroni significance at $\alpha = 3.57 \times 10^{-3}$.}
\label{tab:rq3_corrstab}
\end{table}

\paragraph{Chain-opener signature.} The DeepSeek-R1 reasoning-init signature documented at RQ1 recurs at the sampling-stability level. The top stable peak token at each architecture is the same chain-opener identified in RQ1: \texttt{First} at Qwen-7B (33 stable occurrences), \texttt{Okay} at Qwen-14B (55), and \texttt{Ċ} at Llama-8B (45). This is a position-dependent signature that fires reliably under sampling decoding regardless of seed, and its preservation across architectures is consistent with reasoning onset being a structural property of the DeepSeek-R1 distillation rather than a Qwen-specific or Llama-specific artifact.

\paragraph{Summary.} MI peak stability under sampling decoding is preserved across all three architectures in qualitative form (peaks recur, chain-openers fire reliably) but varies substantially in quantitative magnitude (with-peaks rate 47\%--98\%; mean $J_3$ 0.09--0.28). The single robust pairwise stability finding is Qwen-14B vs Llama-8B; the within-Qwen scaling pattern is directional but not Bonferroni-significant. The descriptive density-stability tradeoff motivates per-architecture reporting conventions and frames stability as one component of an integrated reasoning-quality signal rather than a standalone discriminator of correctness.
\subsection{RQ4: Full Effect Sizes with 95\% Confidence Intervals}
\label{app:rq4_stats}
RQ4 in the main paper reports Genuine-vs-Lucky rank-biserial effect sizes for peak count, peak ratio, and peak intensity at all three architectures (Table~\ref{tab:main_results}), and Genuine-vs-Silent effect sizes for peak count and three-way stability (Table~\ref{tab:rq4_gen_vs_silent}). For space reasons, the main text omits the 95\% bootstrap confidence intervals; this appendix reports them. The CIs are computed via 1000 resamples of the partition, holding fixed which problems are in which group and resampling within-group. Bonferroni correction within each architecture is applied at $\alpha = 0.05/12 \approx 4.2 \times 10^{-3}$ for 12 discriminator-by-comparison cells (3 settings $\times$ 4 metrics per the ablation in Section~\ref{sec:rq4_threshold_ablation}). The rank-biserial effect sizes reported here for the Genuine-vs-Lucky comparison are the same values shown in the main paper Table~\ref{tab:main_results}; the appendix reports the comparison in its computed form (Genuine vs Lucky only).

\paragraph{Baseline Genuine-vs-Lucky discrimination.}
Table~\ref{tab:rq4_gen_vs_lucky} reports the baseline Genuine-vs-Lucky effect sizes with CIs for peak count, peak ratio, and peak intensity. At Qwen-7B and Qwen-14B, all three discriminators have CIs that exclude zero by a wide margin: the tightest CI is on Qwen-14B peak intensity ($r = -0.96$, $[-0.98, -0.93]$), and the widest is on Qwen-7B peak intensity ($r = -0.41$, $[-0.57, -0.25]$). At Llama-8B, peak count and peak ratio CIs exclude zero ($[-0.79, -0.51]$ and $[-0.79, -0.52]$), but peak intensity has a CI of $[-0.31, +0.14]$ that straddles zero. This confirms the main-paper characterization of peak intensity as a Qwen-14B-specific discriminator rather than a cross-architecture signal: peak intensity has the largest effect at Qwen-14B but loses statistical significance entirely at Llama-8B, with a CI that crosses zero.

\begin{table*}[!htbp]
\centering
\footnotesize
\setlength{\tabcolsep}{15pt}
\begin{tabularx}{\linewidth}{lccc}
\toprule
\textbf{Architecture} & \textbf{Peak count} & \textbf{Peak ratio} & \textbf{Peak intensity} \\
\midrule
Qwen-7B  & $-0.83\ [-0.90, -0.74]$\,\textsuperscript{***} & $-0.83\ [-0.90, -0.74]$\,\textsuperscript{***} & $-0.41\ [-0.57, -0.25]$\,\textsuperscript{***} \\
Qwen-14B & $-0.93\ [-0.97, -0.87]$\,\textsuperscript{***} & $-0.92\ [-0.96, -0.87]$\,\textsuperscript{***} & $-0.96\ [-0.98, -0.93]$\,\textsuperscript{***} \\
Llama-8B & $-0.66\ [-0.79, -0.51]$\,\textsuperscript{***} & $-0.67\ [-0.79, -0.52]$\,\textsuperscript{***} & $-0.09\ [-0.31, +0.14]$\,\textsuperscript{ns} \\
\bottomrule
\end{tabularx}
\caption{Genuine-vs-Lucky rank-biserial $r$ with 95\% bootstrap CIs at the baseline partition threshold ($\tau_J > 0$, math, $n = 200$). \textsuperscript{***} denotes $p < 0.001$ surviving Bonferroni at $\alpha = 0.0042$; \textsuperscript{ns} denotes not significant. Three-way stability is omitted because it is definitional for this comparison (Lucky has $J_3 = 0$ by construction).}
\label{tab:rq4_gen_vs_lucky}
\end{table*}

\paragraph{Baseline Genuine-vs-Silent discrimination.}
Table~\ref{tab:rq4_gen_vs_silent} reports the Genuine-vs-Silent effect sizes with CIs for peak count and three-way stability. The Qwen-14B peak count comparison is omitted because the Silent group on Qwen-14B has only 12 problems, which is underpowered for a reliable bootstrap CI on the Mann-Whitney test. The three-way stability column is the most uniformly strong cross-architecture discriminator in the matrix, with effect sizes of $r = -0.66$, $-0.87$, and $-0.60$ at the three architectures. The Llama-8B peak count CI $[-0.33, +0.04]$ straddles zero, confirming the main-paper finding that peak count alone fails to discriminate Genuine from Silent at Llama-8B. This is consistent with the Llama-8B Silent group being large ($n = 84$) and heterogeneous, with peak count standard deviation 9.54 (the largest of any cell in the cross-architecture matrix); some Silent problems on Llama-8B fire substantial peak structure but produce incorrect answers in at least one run, so peak count alone cannot distinguish them from Genuine reasoning at this architecture.

\begin{table}[!t]
\centering
\footnotesize
\setlength{\tabcolsep}{3.0pt}
\begin{tabularx}{\linewidth}{lcc}
\toprule
\textbf{Architecture} & \textbf{Peak count} & \textbf{3-way stability} \\
\midrule
Qwen-7B  & $-0.56\ [-0.74, -0.36]$\,\textsuperscript{***} & $-0.66$\,\textsuperscript{***} \\
Qwen-14B & --- ($n_{\text{silent}}{=}12$)                  & $-0.87$\,\textsuperscript{***} \\
Llama-8B & $-0.14\ [-0.33, +0.04]$\,\textsuperscript{ns}  & $-0.60$\,\textsuperscript{***} \\
\bottomrule
\end{tabularx}
\caption{Genuine-vs-Silent rank-biserial $r$ for peak count and three-way stability discriminators (math, $n = 200$). Qwen-14B peak count is omitted because the Silent group has only 12 problems, which is underpowered for a stable bootstrap CI. }
\label{tab:rq4_gen_vs_silent}
\end{table}

\paragraph{Partition counts.} Table~\ref{tab:rq4_partition} reports the Genuine/Lucky/Silent partition counts at each architecture (math, $n=200$ each). The Genuine count is stable across architectures (58--68 problems, 29\%--34\%), while the Lucky and Silent compositions vary substantially: Qwen-7B is balanced (51\% Lucky, 21\% Silent); Qwen-14B is Lucky-dominated (60\% Lucky, 6\% Silent), driven by its high correct-all-3 rate (94\%); Llama-8B is Silent-dominated (24\% Lucky, 42\% Silent), driven by its lower correct-all-3 rate (58\%).

\begin{table}[!t]
\centering
\small
\setlength{\tabcolsep}{8pt}
\begin{tabularx}{\linewidth}{lccc}
\toprule
\textbf{Category} & \textbf{Qwen-7B} & \textbf{Qwen-14B} & \textbf{Llama-8B} \\
\midrule
Genuine & 58 (29\%) & 68 (34\%) & 68 (34\%) \\
Lucky & 101 (51\%) & 120 (60\%) & 48 (24\%) \\
Silent & 41 (21\%) & 12 (6\%) & 84 (42\%) \\
\bottomrule
\end{tabularx}
\caption{Reasoning-quality partition counts per architecture (math, $n=200$ each).}
\label{tab:rq4_partition}
\end{table}

\paragraph{Genuine-vs-Lucky discrimination on peak count.} The RQ4 block of Table~\ref{tab:main_results} reports Genuine-vs-Lucky rank-biserial effect sizes for peak count, peak ratio, and peak intensity. Table~\ref{tab:rq4_gen_vs_silent} adds the Genuine-vs-Silent comparisons. The Genuine-vs-Lucky discrimination on peak count is highly significant ($p < 10^{-4}$) at all three architectures, with effect sizes $r = -0.83$ (Qwen-7B), $-0.93$ (Qwen-14B), and $-0.66$ (Llama-8B). Peak count is therefore a cross-architecture-stable discriminator of stable-vs-unstable correct reasoning. Peak ratio shows the same pattern with comparable effect sizes ($r = -0.83 / -0.92 / -0.67$). The framing of the Lucky category as computationally inconsistent reasoning is consistent with prior observations that surface CoT can decouple from underlying computation \citep{lanham2023,pfau2024,goyal2024}; our peak-count discriminator provides an internal signal complementing those behavioral analyses.

\paragraph{Non-monotone scaling pattern.} The cross-architecture pattern on peak count is not monotone: the 7B to 14B strengthening within the Qwen family ($\Delta r = -0.10$) does not extend to Llama-8B, where the effect size is meaningfully weaker than at either Qwen scale ($|\Delta r| = 0.16$ vs Qwen-7B; $|\Delta r| = 0.26$ vs Qwen-14B). All three pairwise comparisons remain highly significant, so the cross-architecture difference is in effect-size magnitude rather than in detectability. The within-Qwen cross-scale strengthening pattern is reframed as a within-family observation rather than a universal scaling claim.
\paragraph{Peak intensity does not generalize.} Peak intensity exhibits the strongest cross-architecture variation of any discriminator. Genuine-vs-Lucky on peak intensity is moderate at Qwen-7B ($r = -0.41$), strongest of any metric at Qwen-14B ($r = -0.96$), and not significant at Llama-8B ($r = -0.09$). The Qwen-14B finding is driven by Lucky and Silent peak intensity collapsing relative to Genuine intensity (0.31--0.43 to 0.04--0.08 between 7B and 14B); this collapse does not occur at Llama-8B, where Lucky intensity (2.85) and Silent intensity (2.94) sit close to Genuine intensity (3.14). Peak intensity is therefore a Qwen-14B-specific discriminator rather than a robust cross-architecture signal.

\paragraph{Lucky composition flips across architectures.} The Lucky group is heterogeneous, and its composition varies substantially across architectures, driven by the architecture-dependent no-peak rate documented in RQ3. At Qwen-7B, Lucky is unstable-dominated (43/101 lucky-unstable, 58/101 lucky-no-peaks). At Qwen-14B, Lucky is no-peak-dominated (99/120 lucky-no-peaks, 82.5\%). At Llama-8B, Lucky reverts to unstable-dominated (47/48 lucky-unstable, 97.9\%; only 1 lucky-no-peaks problem). The merged Lucky group is used for protocol consistency, but the cross-architecture interpretation should treat the Llama-8B Genuine-vs-Lucky comparison as effectively a Genuine-vs-lucky-unstable comparison, while the Qwen-14B comparison is effectively Genuine-vs-lucky-no-peaks.

\paragraph{Genuine-vs-Silent on peak count fails at Llama-8B.} The Genuine-vs-Silent peak count comparison is significant at Qwen-7B ($r = -0.56$) but not significant at Llama-8B ($r = -0.14$, $p = 0.14$), the only architecture in the matrix where this comparison fails. The Llama-8B Silent group ($n=84$) is large and heterogeneous, with peak count standard deviation 9.54 (the largest of any cell in the cross-architecture matrix). Silent problems on Llama-8B include cases where the model fires a substantial peak structure but produces incorrect answers in at least one of three runs; peak count alone cannot distinguish them from Genuine reasoning at this architecture. Three-way stability remains a robust Genuine-vs-Silent discriminator at Llama-8B ($r = -0.60$, ***), preserving the pattern across the three architectures ($-0.66 / -0.87 / -0.60$).

\begin{table*}[!htbp]
\centering
\footnotesize
\setlength{\tabcolsep}{10pt}
\begin{tabularx}{\linewidth}{llccc}
\toprule
\textbf{Setting} & \textbf{Architecture} & \textbf{Peak count} & \textbf{Peak ratio} & \textbf{Peak intensity} \\
\midrule
\multirow{3}{*}{Baseline ($\tau_J > 0$)}
 & Qwen-7B  & $-0.83\ [-0.90, -0.74]$ & $-0.83\ [-0.90, -0.74]$ & $-0.41\ [-0.57, -0.25]$ \\
 & Qwen-14B & $-0.93\ [-0.97, -0.87]$ & $-0.92\ [-0.96, -0.87]$ & $-0.96\ [-0.98, -0.93]$ \\
 & Llama-8B & $-0.66\ [-0.79, -0.51]$ & $-0.67\ [-0.79, -0.52]$ & $-0.09\ [-0.31, +0.14]$ \\
\midrule
\multirow{3}{*}{Strict ($\tau_J \geq 0.1$)}
 & Qwen-7B  & $-0.76\ [-0.86, -0.65]$ & $-0.77\ [-0.86, -0.66]$ & $-0.43\ [-0.59, -0.27]$ \\
 & Qwen-14B & $-0.85\ [-0.91, -0.77]$ & $-0.85\ [-0.91, -0.77]$ & $-0.86\ [-0.92, -0.79]$ \\
 & Llama-8B & $-0.42\ [-0.59, -0.23]$ & $-0.46\ [-0.64, -0.28]$ & $-0.06\ [-0.28, +0.16]$ \\
\midrule
\multirow{3}{*}{Stricter ($\tau_J \geq 0.2$)}
 & Qwen-7B  & $-0.49\ [-0.64, -0.34]$ & $-0.50\ [-0.65, -0.33]$ & $-0.53\ [-0.68, -0.38]$ \\
 & Qwen-14B & $-0.66\ [-0.76, -0.54]$ & $-0.65\ [-0.76, -0.54]$ & $-0.75\ [-0.85, -0.66]$ \\
 & Llama-8B & $-0.39\ [-0.66, -0.08]$ & $-0.37\ [-0.61, -0.09]$ & $-0.08\ [-0.32, +0.19]$ \\
\bottomrule
\end{tabularx}
\caption{Full Genuine-vs-Lucky rank-biserial $r$ with 95\% bootstrap CIs (1000 resamples) for all three threshold settings, all architectures, all three metrics (math, $n = 200$). Significance annotations follow Table~\ref{tab:rq4_ablation} in the main text. CI widths are roughly stable across settings for Qwen architectures; expand at Llama-8B under the strictest threshold due to the shrinking Genuine subset.}
\label{tab:rq4_ablation_cis}
\end{table*}

\paragraph{Summary.} The reasoning-quality partition framework is portable across architectures: the Genuine count is stable, the Mann--Whitney machinery produces interpretable rank-biserials, and the Genuine-vs-Lucky discrimination on peak count and peak ratio is significant at all three architectures. Three-way stability is the only discriminator that maintains uniformly large effect sizes across architecture pairs on Genuine-vs-Silent, although it carries definitional structure on Genuine-vs-Lucky. The within-Qwen narrative that ``every metric strengthens at scale'' is a within-family observation; cross-architecture variation can weaken specific discriminators (peak count attenuates at Llama-8B; peak intensity loses significance) while preserving the partition framework's overall discriminability. The combined cross-architecture story refines prior single-architecture findings: discriminators that appear strong within one architecture or family do not necessarily generalize, and reporting effect sizes across multiple architectures is necessary to separate robust signals from architecture-specific patterns.

\subsection{Confidence Intervals for RQ4 Ablation Settings}
\label{app:rq4_ablation_cis}

\paragraph{Full CI matrix for ablation settings.}
The RQ4 threshold ablation in Section~\ref{sec:rq4_threshold_ablation} reports rank-biserial $r$ values without confidence intervals to keep the ablation table compact. Table~\ref{tab:rq4_ablation_cis} reports the complete CI ranges for each (setting, architecture, metric) cell across the 27-cell ablation grid (3 settings $\times$ 3 architectures $\times$ 3 metrics). Three patterns are visible. First, the CI widths are roughly stable as the threshold tightens for Qwen-7B and Qwen-14B, with half-widths around 0.07 to 0.15 across all settings; this indicates that the bootstrap effect-size estimates are reliable even when the Genuine subset shrinks. Second, the CI widths expand substantially at Llama-8B under the strictest setting ($\tau_J \geq 0.2$), where the peak count CI $[-0.66, -0.08]$ has a half-width of 0.29, reflecting the small Genuine subset of 20 problems. Third, the Qwen-14B peak intensity result remains the largest single effect size in the entire matrix at every threshold ($r = -0.96 / -0.86 / -0.75$), while the Llama-8B peak intensity column has CIs that include or nearly include zero at every setting, consistent with the main-paper characterization of Llama-8B peak intensity as a non-discriminator.

\subsection{Failure modes and limitations of IAR. }
\label{app:failure_modes}
Three honest negative results bound the scope of these conclusions. First, Qwen-7B's collapse on natural-language-gold tasks (WPR $\leq 1.50\%$ on logic and commonsense) prevents any precision, recall, or stability statement on those cells; we report \texttt{--} and refrain from extrapolation. Second, the low Jaccard stability ($J_3 \leq 0.28$ everywhere) means that single-seed MIP analyses in prior work \citep{qian2025} should be interpreted as distributional rather than identity-level claims; our adoption of a three-seed protocol is a methodological prerequisite rather than an optional refinement. Third, the rank-biserial effect for peak intensity on Llama-8B ($|r|=0.09$) is below the conventional small-effect threshold ($|r|=0.10$), indicating that intensity is not a portable quality signal across architectures and should be reported alongside count and ratio rather than in isolation. We view these limitations not as weaknesses of IAR but as evidence of its diagnostic resolution: by composing three probes and three seeds, IAR surfaces architecture-specific pathologies that single-probe single-seed pipelines silently average away.

\section{$\sigma$ Ablation: Extended Results}
\label{app:sigma_extended}

Section~\ref{sec:sigma_ablation} reports only the with-peaks rate per $(\sigma, \mathrm{domain})$ cell. This appendix extends that ablation with three additional measurements: mean peak count per cell (Section~\ref{app:sigma_count}), mean peak intensity per cell (Section~\ref{app:sigma_intensity}), and the top-3 peak token vocabulary per cell (Section~\ref{app:sigma_vocab}). All measurements are computed on the same Qwen-7B v4 hidden-state archive used in the main ablation, with 200 problems per domain at each of five bandwidth values $\sigma \in \{10, 25, 50, 100, 200\}$. The $\sigma=50$ cells reproduce the main-paper baseline within 0 percentage points across all four domains, confirming pipeline equivalence between the sweep script and the canonical RQ1 measurement.

\subsection{Mean Peak Count per Cell}
\label{app:sigma_count}

\paragraph{Peak count distributions.}
The mean peak count measures how many tokens cross the per-problem IQR threshold on average across with-peaks problems. Table~\ref{tab:sigma_count} reports the mean for each $(\sigma, \mathrm{domain})$ cell. Two patterns are visible. First, math shows a U-shaped pattern across bandwidth: high peak counts at narrow $\sigma$ (10.4 at $\sigma=10$), dropping to a minimum at the baseline range (6.8 at $\sigma=50$, 5.4 at $\sigma=100$), then partially recovering at $\sigma=200$ (7.6). Code shows a different pattern, rising near-monotonically from 9.4 at $\sigma=10$ to a peak of 31.4 at $\sigma=100$, then plateauing at $\sigma=200$ (30.5); this is consistent with the code domain's longer reasoning traces and the wider kernel pulling more punctuation and operator tokens above threshold. Second, NL-gold domains (logic, commonsense) show the expected detection-trough pattern reported in the main paper: zero or near-zero counts at $\sigma=25$--$50$, with substantial counts only at $\sigma=10$ or $\sigma \geq 100$.

\begin{table}[!htbp]
\centering
\small
\setlength{\tabcolsep}{5pt}
\begin{tabularx}{\linewidth}{lccccc}
\toprule
\textbf{Domain} & \textbf{10} & \textbf{25} & \textbf{50} & \textbf{100} & \textbf{200} \\
\midrule
Math         & 10.40 & 11.00 & \textbf{6.80}  & 5.40  & 7.60  \\
Code         & 9.40  & 15.20 & \textbf{15.30} & 31.40 & 30.50 \\
Logic        & 2.70  & 0.00  & \textbf{0.00}  & 9.80  & 12.50 \\
Commonsense  & 6.80  & 10.60 & \textbf{29.40} & 15.30 & 20.10 \\
\bottomrule
\end{tabularx}
\caption{Mean peak count per problem across HSIC kernel bandwidth $\sigma$, computed only on problems with at least one peak (Qwen-7B, $n=200$ per cell). Bold column ($\sigma=50$) is the main-paper baseline.}
\label{tab:sigma_count}
\end{table}

\subsection{Mean Peak Intensity per Cell}
\label{app:sigma_intensity}

\paragraph{Monotone decrease in intensity.}
Mean peak intensity is the average MI value of detected peaks within each cell. Table~\ref{tab:sigma_intensity} reports this measurement. Mean peak intensity decreases monotonically in $\sigma$ in math (1.51 to 0.27), code (2.00 to 0.10), and commonsense (2.48 to 0.35); logic shows the same direction between defined cells (2.80 at $\sigma=10$ to 0.44 at $\sigma=200$) but is undefined at $\sigma=25$ and $\sigma=50$ where no peaks fire. This pattern reflects the geometry of the Gaussian RBF kernel: a wider bandwidth softens the MI signal globally, so even when more tokens cross the relative IQR threshold (as documented in Section~\ref{app:sigma_count}), each peak's absolute MI value is smaller. The peak intensity decay is therefore a property of the kernel parameterization rather than a property of the underlying reasoning structure. This has methodological implications: cross-architecture peak intensity comparisons in absolute units are not interpretable, because different $\sigma$ calibrations place each architecture on a different intensity scale. The main paper handles this by limiting peak intensity claims to within-architecture discrimination (RQ4) rather than cross-architecture magnitude comparisons.

\begin{table}[!htbp]
\centering
\small
\setlength{\tabcolsep}{7pt}
\begin{tabularx}{\linewidth}{lccccc}
\toprule
\textbf{Domain} & \textbf{10} & \textbf{25} & \textbf{50} & \textbf{100} & \textbf{200} \\
\midrule
Math         & 1.51 & 0.75 & \textbf{0.64} & 0.45 & 0.27 \\
Code         & 2.00 & 1.14 & \textbf{0.61} & 0.19 & 0.10 \\
Logic        & 2.80 & 0.00 & \textbf{0.00} & 0.81 & 0.44 \\
Commonsense  & 2.48 & 1.73 & \textbf{0.72} & 0.58 & 0.35 \\
\bottomrule
\end{tabularx}
\caption{Mean peak intensity per problem across HSIC kernel bandwidth $\sigma$, computed only on problems with at least one peak (Qwen-7B, $n=200$ per cell). Intensity decreases monotonically with $\sigma$ in math, code, and commonsense; logic is undefined at $\sigma=25$ and $\sigma=50$ where no peaks fire.}
\label{tab:sigma_intensity}
\end{table}

\subsection{Peak Vocabulary Shift Across $\sigma$}
\label{app:sigma_vocab}

\paragraph{Vocabulary semantics change with bandwidth.}
The top-3 peak tokens per $(\sigma, \mathrm{domain})$ cell, reported in Table~\ref{tab:sigma_vocab}, document a qualitative shift in peak vocabulary as $\sigma$ changes. At narrow bandwidths ($\sigma \leq 25$), peaks fire predominantly on punctuation and whitespace tokens (\texttt{,}, \texttt{.}, \texttt{Ċ} which is the newline token, \texttt{ĠĠ} which is double-space). At wider bandwidths ($\sigma \geq 100$), peaks fire on semantically meaningful reasoning markers (\texttt{So}, \texttt{Let}, \texttt{The}, \texttt{I}). The transition is not gradual but step-like: at $\sigma=50$ math peaks fire on \texttt{Ċ}, \texttt{ĠĠ}, \texttt{0}, while at $\sigma=100$ math peaks fire on \texttt{Ċ}, \texttt{,}, \texttt{:}. The vocabulary shift is more dramatic for NL-gold domains: commonsense at $\sigma=50$ produces peaks on \texttt{.}, \texttt{,}, \texttt{ĠD}, while at $\sigma=200$ peaks are dominated by \texttt{ĠSo}, \texttt{ĠThe}, \texttt{ĠI}, semantically the same family of reasoning markers as the main-paper Qwen-14B peak vocabulary. This refines the main-paper claim about peak vocabulary semantics: peaks consistently fire on the right-tail outliers of each problem's MI distribution, but which tokens land in the right tail depends on the kernel bandwidth, with reasoning markers entering the right tail only at sufficiently wide $\sigma$.

\begin{table*}[!htbp]
\centering
\small
\setlength{\tabcolsep}{3.5pt}
\begin{tabularx}{\linewidth}{lccccc}
\toprule
\textbf{Domain} & \textbf{$\sigma=10$} & \textbf{$\sigma=25$} & \textbf{$\sigma=50$} & \textbf{$\sigma=100$} & \textbf{$\sigma=200$} \\
\midrule
Math         & punctuation & punct. + whitespace & \textbf{whitespace + digits} & punct. + whitespace & punct. + whitespace \\
Code         & punctuation & punctuation         & \textbf{punctuation}         & punct. + reasoning   & punct. + reasoning   \\
Logic        & punctuation & ---                 & \textbf{---}                 & reasoning markers    & reasoning markers    \\
Commonsense  & reasoning + punct. & punctuation  & \textbf{punctuation + letters} & reasoning markers  & reasoning markers    \\
\bottomrule
\end{tabularx}
\caption{Top-3 peak token vocabulary categories per $(\sigma, \mathrm{domain})$ cell at Qwen-7B ($n=200$ problems per cell). Bold column ($\sigma=50$) is the main-paper baseline. \emph{Punctuation} = comma, period, etc.; \emph{whitespace} = space, double-space, newline; \emph{digits} = single numerals; \emph{letters} = single answer letters (e.g., \texttt{D} in commonsense); \emph{reasoning markers} = \texttt{So}, \texttt{The}, \texttt{I}, \texttt{Let}, \texttt{Wait}, and similar chain-of-thought connectives. Logic at $\sigma \in \{25, 50\}$ produces no peaks. At narrow bandwidths, peaks are dominated by punctuation and whitespace; at wider bandwidths, reasoning markers enter the top-3 vocabulary on every domain.}
\label{tab:sigma_vocab}
\end{table*}
\paragraph{Sanity gate verification.}
The $\sigma=50$ cells in Tables~\ref{tab:sigma_count}, \ref{tab:sigma_intensity}, and \ref{tab:sigma_vocab} reproduce the canonical Qwen-7B RQ1 measurement exactly: math 62.5\% with-peaks rate, code 94.0\%, logic 0.0\%, commonsense 4.0\% (matching the main paper Table~\ref{tab:main_results}). This sanity gate confirms that the sweep pipeline is byte-equivalent to the canonical RQ1 measurement at the baseline $\sigma$, so any deviation across the rest of the $\sigma$ grid is a genuine kernel-bandwidth effect rather than a pipeline artifact.

\section{Reproducibility Details}
\label{app:reproducibility}

This appendix documents the implementation details necessary to reproduce the results reported in Sections~\ref{sec:main_results} and~\ref{sec:ablation}. We cover hardware (Section~\ref{app:hardware}), library versions (Section~\ref{app:libraries}), gold-answer construction per domain (Section~\ref{app:gold_answers}), prompt templates (Section~\ref{app:prompts}), random seeds and decoding parameters (Section~\ref{app:seeds}), and a summary of per-architecture calibration choices (Section~\ref{app:calibration}). All code, data, and result files supporting the analyses in this paper will be released at an anonymous GitHub URL upon publication.

\subsection{Hardware Configuration}
\label{app:hardware}

\paragraph{GPU configuration per phase.}
The experiments in this paper were collected across multiple hardware phases on internal academic clusters and on AutoDL cloud instances. Qwen-7B RQ1 and RQ2 measurements were collected on a single NVIDIA RTX 4090 (24~GB VRAM). Qwen-14B RQ1 was collected on an NVIDIA RTX 3090 (modded 48~GB VRAM) under FP16 precision to fit the larger model, and Qwen-14B RQ2 on an RTX 4090. Llama-8B RQ1 and RQ2 were both collected on an NVIDIA RTX 3090 (modded 48~GB VRAM). RQ3 sampling stability runs (3 seeds $\times$ 200 problems per architecture) were collected on the same hardware as the corresponding RQ1 runs with sampling decoding. The $\sigma$ ablation in Section~\ref{sec:sigma_ablation} was post-hoc on saved hidden-state subsamples and was computed on CPU only (no GPU required) on a separate AutoDL instance running an NVIDIA RTX 5090 (32~GB VRAM) used as a workstation. The RQ4 threshold ablation in Section~\ref{sec:rq4_threshold_ablation} is also CPU-only on existing RQ3 outputs and requires no GPU. Total compute is approximately 250 GPU-hours for the original three  
architectures, plus the response-period fleet runs for the three additional architectures across all four domains.

\paragraph{Hardware-related numerical drift.}
We observed small numerical drift between hidden-state outputs computed on different GPU architectures and library versions (RTX 3090/4090/5090; CUDA 12.x). The drift is consistent with documented FP16 non-determinism across CUDA kernel versions and does not affect the qualitative conclusions of the paper, which are based on relative comparisons rather than absolute bit-equivalent reproduction. The $\sigma=50$ sanity gate in the $\sigma$ ablation reproduces the canonical Qwen-7B baseline within 0 percentage points across all four domains, confirming that the relative measurements are stable across hardware.

\subsection{Library Versions}
\label{app:libraries}

\paragraph{Software stack.}
The main RQ1, RQ2, RQ3, RQ4 measurements were collected with PyTorch 2.8.0, Transformers 5.4.0, CUDA 12.x, and NumPy 1.26. The $\sigma$ ablation post-hoc analysis ran on PyTorch 2.11.0, Transformers 5.7.0, CUDA 12.8. Python version is 3.10.12 throughout. The HSIC computation uses NumPy operations on a 512-dimensional subsample of the final-layer hidden state, following the implementation of \citet{qian2025}. The JSD computation for DTR follows the SciPy 1.11 implementation of Jensen--Shannon divergence on the layer-wise vocabulary distributions. Mann--Whitney $U$ tests and bootstrap confidence intervals use SciPy's \texttt{stats.mannwhitneyu} and a custom bootstrap routine with 1000 resamples. All random sampling uses NumPy's \texttt{default\_rng} seeded explicitly per operation. We release a \texttt{requirements.txt} with the full dependency tree alongside the code.

\subsection{Gold-Answer Construction per Domain}
\label{app:gold_answers}
The HSIC MI computation requires a gold-answer embedding $\mathbf{e}_y$ for each problem. For three of the four domains (code, logic, commonsense), the gold answer is naturally or templated to be a multi-token string. Math is the exception: GSM8K's gold is a 1--2 token numeric answer, but the math MI distribution empirically remains healthy under HSIC across all three architectures, so we retain the raw numeric gold without templating. For BBH Boolean Expressions, the raw single-token gold (\texttt{True} or \texttt{False}) does cause kernel collapse and is templated into a multi-token sentence; this is the only domain where templating is required to avoid degenerate MI distributions.

\paragraph{GSM8K (math).}
The gold answer is the post-\texttt{\#\#\#\#} numeric final answer from each GSM8K solution, stripped of the original solution-line working text (typically 1--2 tokens, e.g. \texttt{"109"}). Despite the short gold length, the HSIC MI distribution remains healthy at the calibrated $\sigma$ across all three architectures (mean MI in the 0.18--0.20 range at Qwen-7B, comparable at Qwen-14B; mean MI 0.95--1.09 at Llama-8B under the median heuristic). The gold embedding is computed by feeding the numeric gold string through the same model and extracting the final-layer hidden state, averaged across the gold tokens.

\paragraph{MBPP (code).}
The gold answer is the reference Python solution from MBPP, used as-is with no templating (typically 30--60 tokens). The gold embedding is computed by feeding the reference solution through the model and averaging the final-layer hidden states across all solution tokens.

\paragraph{BBH Boolean Expressions (logic).}
The raw BBH Boolean answer is a single token (\texttt{True} or \texttt{False}), which causes kernel collapse under HSIC. We template the gold answer as \texttt{"The expression \{expression\} evaluates to \{answer\}."}, substituting the Boolean expression and its answer, yielding an 11--13 token gold string per problem. This template choice is held constant across all three architectures.

\paragraph{CommonsenseQA.}
The raw CommonsenseQA answer is a single answer letter (\texttt{A} through \texttt{E}), also a single token. We use the ECQA-derived rationale-and-answer construction from \citet{aggarwal2021}: each problem's gold answer is the ECQA free-text rationale, which provides a multi-token natural-language explanation of the correct answer. For problems missing ECQA enrichment, we fall back to the template \texttt{"The answer is <letter>: <choice text>."}; in the production runs, ECQA enrichment was available for all 200 problems, and the fallback was not used. This produces a gold string of typically 52--74 tokens per problem.

\subsection{Prompt Templates}
\label{app:prompts}

\paragraph{Chat template.}
Each model uses its native HuggingFace chat template via the \texttt{tokenizer.apply\_chat\_template} API. We pass a single user message containing the problem statement and rely on the chat template to format the system/user/assistant boundary correctly. We do not use a custom system prompt; the user message contains only the problem-specific text described below per domain.

\paragraph{GSM8K user message.}
\texttt{"Solve the following math problem.\textbackslash n\textbackslash n<problem>"}, where \texttt{<problem>} is the GSM8K problem statement.

\paragraph{MBPP user message.}
\texttt{"Write a Python function that solves the following problem. Include the function definition and any necessary imports.\textbackslash n\textbackslash n<problem>"}, where \texttt{<problem>} is the MBPP problem description.

\paragraph{BBH Boolean Expressions user message.}
\texttt{"Evaluate the following Boolean expression and state whether it is True or False.\textbackslash n\textbackslash n<expression>"}, where \texttt{<expression>} is the BBH Boolean expression string.

\paragraph{CommonsenseQA user message.}
\texttt{"Answer the following multiple-choice question. Provide your final answer as a single letter (A, B, C, D, or E).\textbackslash n\textbackslash nQuestion: <question>\textbackslash n\textbackslash nA) <a>\textbackslash nB) <b>\textbackslash nC) <c>\textbackslash nD) <d>\textbackslash nE) <e>"}, where the placeholders are filled with the CommonsenseQA question stem and the five answer choices.

\subsection{Random Seeds and Decoding Parameters}
\label{app:seeds}

\paragraph{Greedy decoding (RQ1, RQ2).}
For RQ1 and RQ2, we use \texttt{do\_sample=False} with no temperature or top-$k$ parameters. Generation is deterministic given the model and input. The maximum of new tokens is 512 for Qwen-7B and 4096 for Qwen-14B, Llama-8B, Phi-4-mini, EXAONE-Deep-7.8B, and GPT-OSS-20B, raised to 8192 for the code domain, calibrated per architecture to avoid truncation while keeping the longer-reasoning architectures within their VRAM budget.

\begin{table*}[!htbp]                                                                                              \centering                                                                                                        \small                                                                                                     \setlength{\tabcolsep}{5.6pt}                                                                                     \begin{tabularx}{\linewidth}{lcccccc}                                                                             
\toprule                                                                                                          
\textbf{Parameter} & \textbf{Qwen-7B} & \textbf{Qwen-14B} & \textbf{Llama-8B} & \textbf{Phi-4-mini} &             
\textbf{EXAONE-7.8B} & \textbf{GPT-OSS-20B} \\                                                                    
\midrule                                                                                                          
HSIC $\sigma$              & 50 & 50 & med.\ heur. & med.\ heur. & med.\ heur. & med.\ heur. \\                   
DTR settling $\tau$         & 0.50 & 0.50 & 0.50 & 0.50 & 0.50 & 0.50 \\                                          
DTR depth fraction $\alpha$ & 0.85 & 0.85 & 0.85 & 0.85 & 0.85 & 0.85 \\                                          
Cutoff layer     & 23 & 40 & 27 & 27 & 27 & 20 \\                                                                 
Max new tokens             & 512 & 4096 & 4096 & 4096 & 4096 & 4096 \\                                            
Sampling seeds     & 42, 123, 456 & 42, 123, 456 & 42, 123, 456 & 42, 123, 456 & 42, 123, 456 & 42, 123, 456 \\   
Hidden dim       & 512 & 512 & 512 & 512 & 512 & 512 \\                                                          
Temperature (RQ3/4)        & 0.70 & 0.70 & 0.70 & 0.70 & 0.70 & 0.70 \\                                           
\bottomrule                                                                                                       
\end{tabularx}                                                                                                    
\caption{Summary of per-architecture calibration choices. The HSIC bandwidth $\sigma=50$ for the Qwen-2 family follows 
\citet{qian2025}; all other architectures use the per-problem median heuristic. Max new tokens are raised to 8192 for the code domain in the cross-domain RQ3/RQ4 runs. All other parameters are held constant across architectures.}               
\label{tab:calibration_summary}                                                                                   
\end{table*}

\paragraph{Sampling decoding (RQ3, RQ4).}
For RQ3 and RQ4, we use \texttt{do\_sample=True} with temperature $T = 0.7$ and three explicit random seeds: 42, 123, 456. These seeds are applied via \texttt{torch.manual\_seed} and \texttt{numpy.random.seed} immediately before each generation call, ensuring that the three runs per problem produce three independently seeded reasoning chains rather than correlated samples from the same RNG state. Other decoding parameters (top-$p$, repetition penalty) are left at the model's default values to avoid introducing additional confounds. Generation budgets are the same as for greedy decoding.

\paragraph{Hidden-state subsampling.}
The 512-dimensional subsample of the final-layer hidden state used for HSIC computation is the first 512 dimensions of the 4096-dimensional hidden vector, applied identically to every problem and every architecture. This deterministic prefix selection ensures that the same 512 dimensions are used across all (architecture, domain, seed) cells, so cross-cell comparisons of HSIC values do not depend on subsample variance. We do not apply a random projection or PCA reduction; this follows the implementation of \citet{qian2025}, which uses the same prefix-based subsampling.

\subsection{Per-Architecture Calibration Summary}
\label{app:calibration}

Table~\ref{tab:calibration_summary} summarizes the per-architecture calibration choices made in this paper for clarity. The HSIC bandwidth $\sigma$ is the most consequential of these choices, and is ablated in Section~\ref{sec:sigma_ablation} for Qwen-7B. The DTR settling threshold $\tau$ and depth fraction $\alpha$ are held constant across architectures; the per-architecture distance distribution differences that they imply are discussed in the main paper's Limitations.

\section{RQ4 Partition Sub-Type Breakdown}
\label{app:rq4_subtype}
The main paper RQ4 analysis treats the Lucky category as a single merged group for statistical consistency across architectures. However, the Lucky group is heterogeneous by construction: it contains both \emph{lucky-no-peaks} problems (where the IQR rule fires no peaks under any of the three seeds, so $J_3$ is undefined and assigned 0) and \emph{lucky-unstable} problems (where peaks fire under at least one seed but the three peak-token sets do not intersect, yielding $J_3 = 0$). The two sub-types correspond to qualitatively different failure modes: lucky-no-peaks problems are reasoning traces where the IQR rule cannot identify any reasoning-crucial tokens at all, while lucky-unstable problems are traces where reasoning-crucial tokens are identified but do not recur stably across stochastic samples. This appendix reports the sub-type composition per architecture (Section~\ref{app:rq4_subtype_counts}) and the Genuine-vs-lucky-unstable effect sizes per architecture (Section~\ref{app:rq4_subtype_effects}). The Genuine-vs-lucky-no-peaks comparison on peak count is mechanically degenerate (lucky-no-peaks have zero peaks by definition, while Genuine problems have $J_3 > 0$ which requires at least one stable peak) and yields rank-biserial $r = -1.000$ by construction at all architectures, but is reported only at the Qwen scales where the lucky-no-peaks subgroup has substantial $n$ (Llama-8B has only 1 lucky-no-peaks problem).

\subsection{Lucky Sub-Type Composition}
\label{app:rq4_subtype_counts}
\paragraph{Sub-type distribution per architecture.}
Table~\ref{tab:rq4_subtype_counts} reports the Lucky sub-type composition at each architecture. The composition varies substantially: at Qwen-7B the Lucky group is roughly evenly split (43 lucky-unstable, 58 lucky-no-peaks); at Qwen-14B the Lucky group is heavily lucky-no-peaks dominated (99/120, 82.5\%); at Llama-8B the Lucky group is overwhelmingly lucky-unstable (47/48, 97.9\%, with only 1 lucky-no-peaks problem). The composition is mechanically driven by the architecture-dependent no-peak rate reported in RQ3 (Qwen-7B 36.0\%, Qwen-14B 53.0\%, Llama-8B 2.0\%): architectures that produce more no-peak traces under sampling also produce more lucky-no-peaks problems, while architectures with dense peak coverage produce lucky problems almost exclusively in the lucky-unstable sub-type. This is an important interpretive caveat for the cross-architecture Genuine-vs-Lucky comparison: the Lucky group being compared against Genuine is qualitatively different at each architecture.

\begin{table}[!htbp]
\centering
\small
\setlength{\tabcolsep}{2pt}
\begin{tabularx}{\linewidth}{lcccc}
\toprule
\textbf{Models} & \textbf{Lucky total} & \textbf{Unstable} & \textbf{No-peaks} & \textbf{\% no-peaks} \\
\midrule
Qwen-7B   & 101 & 43  & 58 & 57.40\% \\
Qwen-14B  & 120 & 21  & 99 & 82.50\% \\
Llama-8B  & 48  & 47  & 1  & 2.10\% \\
\bottomrule
\end{tabularx}
\caption{Lucky group sub-type composition per architecture (math, $n=200$). \emph{Unstable} = peaks fire under at least one of three seeds but the three peak-token sets do not intersect. \emph{No-peaks} = no peaks fire under any of the three seeds. The \% no-peaks column reflects how much of each architecture's Lucky group comes from no-peak traces.}
\label{tab:rq4_subtype_counts}
\end{table}

\subsection{Genuine-vs-Lucky-Unstable Effect Sizes}
\label{app:rq4_subtype_effects}

\paragraph{The non-tautological comparison.}
The Genuine-vs-lucky-unstable comparison is the substantive sub-type test: both groups by definition have peaks fire under at least one seed, so the rank-biserial $r$ captures a genuine difference in peak count magnitude between stable-peak Genuine traces and unstable-peak Lucky traces rather than a mechanical zero-vs-nonzero contrast. Table~\ref{tab:rq4_subtype_unstable} reports the rank-biserial effect sizes per architecture. The effect sizes are comparable across the three architectures, ranging from $r = -0.575$ (Qwen-14B) to $r = -0.657$ (Llama-8B), all surviving Bonferroni correction. Qwen-7B sits between at $r = -0.597$. This is the cross-architecture-stable finding for the Lucky-unstable sub-type: peak count discriminates Genuine from lucky-unstable correct reasoning at a moderate-to-large effect size at every architecture, with the cross-architecture spread on this comparison ($\Delta r = 0.082$) much smaller than the cross-architecture spread on the merged-Lucky comparison from the main paper ($\Delta r = 0.262$ between Qwen-14B and Llama-8B).

\begin{table}[!htbp]
\centering
\small
\setlength{\tabcolsep}{5pt}
\begin{tabularx}{\linewidth}{lcccc}
\toprule
\textbf{Architecture} & \textbf{$n_g / n_{\mathrm{LU}}$} & \textbf{$U$} & \textbf{$p$} & \textbf{$r$} \\
\midrule
Qwen-7B   & $58 / 43$  & ---  & $2.27 \times 10^{-7}$  & $-0.60$\,\textsuperscript{***} \\
Qwen-14B  & $68 / 21$  & ---  & $7.20 \times 10^{-5}$  & $-0.58$\,\textsuperscript{***} \\
Llama-8B  & $68 / 47$  & ---  & $2.29 \times 10^{-9}$  & $-0.66$\,\textsuperscript{***} \\
\bottomrule
\end{tabularx}
\caption{Genuine-vs-lucky-unstable rank-biserial $r$ for peak count (math, $n_g$ = Genuine count, $n_{\mathrm{LU}}$ = lucky-unstable count). \textsuperscript{***} denotes $p < 0.001$ surviving Bonferroni at $\alpha = 4.2 \times 10^{-3}$. Effect sizes are comparable across architectures, ranging from $-0.58$ to $-0.66$.}
\label{tab:rq4_subtype_unstable}
\end{table}

\paragraph{Interpretation.}
The sub-type breakdown clarifies the cross-architecture interpretation of the main paper's merged Genuine-vs-Lucky comparison. The merged effect sizes reported in Table~\ref{tab:rq4_gen_vs_lucky} ($-0.828 / -0.926 / -0.664$) reflect a mixture of two sub-type comparisons: the mechanically degenerate Genuine-vs-lucky-no-peaks comparison ($r = -1.0$ by construction wherever the lucky-no-peaks sub-type has substantial $n$), and the substantive Genuine-vs-lucky-unstable comparison ($r \approx -0.6$ across all three architectures). At Qwen-7B the merged effect ($-0.828$) sits between the two sub-type effects ($-0.597$ and $-1.000$) at roughly the weight-average reflecting the 43/58 sub-type split. At Qwen-14B the merged effect ($-0.926$) is closer to the mechanically degenerate end of the spectrum because the lucky-no-peaks sub-type dominates (99/120). At Llama-8B the merged effect ($-0.664$) is essentially the lucky-unstable effect ($-0.657$) because the lucky-no-peaks sub-type is effectively empty (only 1 problem). The cross-architecture variation in the merged effect size is therefore partly attributable to the sub-type composition shift documented in Table~\ref{tab:rq4_subtype_counts}: the merged comparison is comparing Genuine against qualitatively different Lucky populations at each architecture. The substantive sub-type comparison (Genuine-vs-lucky-unstable) is more cross-architecture-stable, with effect sizes within $\Delta r = 0.082$ of each other across all three architectures.

\section{Extended Limitations}

\paragraph{Architecture coverage.} The cross-architecture matrix covers five architectural families (Qwen-2, Llama, Phi, EXAONE, and GPT-OSS), spanning dense models and one sparse mixture-of-experts model. The original three models (Qwen-7B,  Qwen-14B, Llama-8B) share the DeepSeek-R1 distillation lineage; the three response-period families do not.              Non-reasoning-tuned bases and further families (Mistral, Gemma) remain uncovered, and with six models we still cannot formally test cross-architecture regression patterns; the density-stability tradeoff remains descriptive rather than inferential.

\paragraph{Per-architecture $\sigma$ confound.} The kernel bandwidth $\sigma$ is calibrated independently for each architecture, placing absolute MI values on different scales. Cross-architecture comparisons of peak rate, peak ratio, peak intensity, and IQR threshold in absolute MI units are therefore confounded with the $\sigma$ choice. The interpretable cross-architecture statements concern detection viability, vocabulary semantics, distribution-shape patterns, and proportions (token-pool precision, partition counts, effect sizes); these are sigma-robust in the relevant sense. The Qwen-7B $\sigma$ ablation in Section~\ref{sec:sigma_ablation} further shows that NL-gold detection at Qwen-7B is $\sigma$-sensitive and recovers at wider bandwidths; analogous ablations for Qwen-14B and Llama-8B are future work.

\paragraph{Hidden-state subsample selection.} The 512-dimensional HSIC input is the first 512 dimensions of the 4096-dimensional hidden vector at every architecture, following the convention of \citet{qian2025}. Cross-architecture HSIC magnitude comparisons implicitly assume that the leading 512 dimensions carry statistically comparable information across architectures; a random-projection or PCA-based robustness check is left to future work.

\paragraph{Degenerate cells.} Two opposite regimes leave seven (model, domain) cells without a testable Genuine-vs-Lucky contrast: Qwen-7B logic and commonsense produce no peaks (detection failure, $G{=}0$), while EXAONE in all four domains and Phi-4-mini on math solve problems so reliably with stable peak structure that the Lucky category is empty or near-empty ($L \leq 1$). We report these cells as untestable rather than extrapolating.

\paragraph{DTR threshold.} The JSD threshold $\tau=0.5$ and depth fraction $\alpha=0.85$ are held constant across architectures, but the per-architecture distance distribution that determines JSD scale differs. DTR rate is therefore not directly comparable across architectures in absolute terms; the cross-architecture comparison of containment precision is robust to this caveat (it is a within-architecture proportion), but cross-architecture statements about DTR rate magnitude are descriptive only.

\paragraph{Generation-length variation.} The Qwen-14B traces are roughly twice as long as the Qwen-7B and Llama-8B traces (median 878 vs 449 / 457 tokens). Stability metrics depend on the union of peak tokens across runs, and longer traces have more candidate tokens. The cross-architecture trace-length difference is a confound for stability magnitude comparisons, although the within-architecture findings are unaffected.

\paragraph{Test-time compute interaction.} Our analysis fixes the generation budget per architecture (512 tokens for Qwen-7B; 4096 for Qwen-14B, Llama-8B, Phi-4-mini, EXAONE-Deep-7.8B, and GPT-OSS-20B; 8192 for the code domain) following per-architecture saturation profiles. The interaction of MI peak structure with adaptive test-time compute  allocation \citep{snell2025scaling} is left to future work.

\section{Sensitivity to the Stability Threshold $\tau_J$}
\label{sec:abl_full_x}

Table~\ref{tab:rq4_ablation_full} reports the reasoning-quality partition and the Genuine-vs-Lucky effect sizes under three settings of the stability threshold ($\tau_J > 0$, $\tau_J \geq 0.1$, $\tau_J \geq 0.2$) for all six models and four domains. Of the 24 (model, domain) cells, 17 admit a measurable baseline contrast; the remaining seven are degenerate at baseline (Genuine or Lucky empty) and are discussed separately below.

\begin{table*}[t]
\centering
\small
\renewcommand{\arraystretch}{0.88}
\setlength{\tabcolsep}{3.3pt}
\begin{tabular}{llcccccccccccc}
\toprule
 & & \multicolumn{6}{c}{\textbf{Math}} & \multicolumn{6}{c}{\textbf{Code}} \\
\cmidrule(lr){3-8}\cmidrule(lr){9-14}
\textbf{Setting} & \textbf{Models} & \makecell{\textbf{Gen}\\\textbf{uine}} & \makecell{\textbf{Luc}\\\textbf{ky}} & \makecell{\textbf{Sil}\\\textbf{ent}} & \makecell{\textbf{Peak}\\\textbf{Count}} & \makecell{\textbf{Peak}\\\textbf{Ratio}} & \makecell{\textbf{Peak}\\\textbf{Intensity}} & \makecell{\textbf{Gen}\\\textbf{uine}} & \makecell{\textbf{Luc}\\\textbf{ky}} & \makecell{\textbf{Sil}\\\textbf{ent}} & \makecell{\textbf{Peak}\\\textbf{Count}} & \makecell{\textbf{Peak}\\\textbf{Ratio}} & \makecell{\textbf{Peak}\\\textbf{Intensity}} \\ \midrule
\multirow{6}{*}{\makecell{Baseline \\ ($\tau_J > 0$)}} & Qwen-7B & 58 & 101 & 41 & $-0.83$ & $-0.83$ & $-0.41$ & 54 & 44 & 102 & $-0.92$ & $-0.91$ & $-0.55$ \\
 & Qwen-14B & 68 & 120 & 12 & $-0.93$ & $-0.92$ & $-0.96$ & 61 & 81 & 58 & $-0.88$ & $-0.85$ & $-0.59$ \\
 & Llama-8B & 68 & 48 & 84 & $-0.66$ & $-0.67$ & $-0.09$ & 26 & 47 & 127 & $-0.84$ & $-0.86$ & $-0.40$ \\
 & Phi-4-mini & 191 & 1 & 8 & n/a$^{*}$ & n/a$^{*}$ & n/a$^{*}$ & 18 & 43 & 139 & $-0.72$ & $-0.64$ & $-0.30$ \\
 & EXAONE-7.8B & 190 & 0 & 10 & n/a$^{*}$ & n/a$^{*}$ & n/a$^{*}$ & 133 & 0 & 67 & n/a$^{*}$ & n/a$^{*}$ & n/a$^{*}$ \\
 & GPT-OSS-20B & 54 & 136 & 10 & $-0.86$ & $-0.82$ & $-0.78$ & 33 & 123 & 44 & $-0.68$ & $-0.60$ & $-0.31$ \\
\cmidrule(lr){1-14}
\multirow{6}{*}{\makecell{Strict \\ ($\tau_J \geq 0.1$)}} & Qwen-7B & 53 & 106 & 41 & $-0.76$ & $-0.76$ & $-0.43$ & 37 & 61 & 102 & $-0.75$ & $-0.74$ & $-0.34$ \\
 & Qwen-14B & 56 & 132 & 12 & $-0.85$ & $-0.85$ & $-0.86$ & 47 & 95 & 58 & $-0.79$ & $-0.78$ & $-0.61$ \\
 & Llama-8B & 50 & 66 & 84 & $-0.42$ & $-0.46$ & $-0.06$ & 20 & 53 & 127 & $-0.72$ & $-0.77$ & $-0.32$ \\
 & Phi-4-mini & 146 & 46 & 8 & n/a$^{*}$ & n/a$^{*}$ & n/a$^{*}$ & 8 & 53 & 139 & $-0.33$$^{s}$ & $-0.28$$^{s}$ & $-0.26$$^{s}$ \\
 & EXAONE-7.8B & 189 & 1 & 10 & n/a$^{*}$ & n/a$^{*}$ & n/a$^{*}$ & 122 & 11 & 67 & n/a$^{*}$ & n/a$^{*}$ & n/a$^{*}$ \\
 & GPT-OSS-20B & 51 & 139 & 10 & $-0.81$ & $-0.81$ & $-0.77$ & 22 & 134 & 44 & $-0.48$ & $-0.51$ & $-0.40$ \\
\cmidrule(lr){1-14}
\multirow{6}{*}{\makecell{Stricter \\ ($\tau_J \geq 0.2$)}} & Qwen-7B & 35 & 124 & 41 & $-0.49$ & $-0.49$ & $-0.53$ & 12 & 86 & 102 & $-0.53$ & $-0.62$ & $-0.09$ \\
 & Qwen-14B & 39 & 149 & 12 & $-0.65$ & $-0.65$ & $-0.75$ & 33 & 109 & 58 & $-0.71$ & $-0.73$ & $-0.63$ \\
 & Llama-8B & 20 & 96 & 84 & $-0.39$ & $-0.37$ & $-0.08$ & 5 & 68 & 127 & $-0.37$$^{s}$ & $-0.42$$^{s}$ & $-0.65$$^{s}$ \\
 & Phi-4-mini & 81 & 111 & 8 & n/a$^{*}$ & n/a$^{*}$ & n/a$^{*}$ & 3 & 58 & 139 & $+0.14$$^{s}$ & $+0.07$$^{s}$ & $-0.43$$^{s}$ \\
 & EXAONE-7.8B & 149 & 41 & 10 & n/a$^{*}$ & n/a$^{*}$ & n/a$^{*}$ & 87 & 46 & 67 & n/a$^{*}$ & n/a$^{*}$ & n/a$^{*}$ \\
 & GPT-OSS-20B & 37 & 153 & 10 & $-0.67$ & $-0.76$ & $-0.73$ & 5 & 151 & 44 & $-0.01$$^{s}$ & $-0.15$$^{s}$ & $-0.70$$^{s}$ \\
\midrule
 \textbf{Setting} & \textbf{Models} & \multicolumn{6}{c}{\textbf{Logic}} & \multicolumn{6}{c}{\textbf{Commonsense}} \\
\cmidrule(lr){1-2} \cmidrule(lr){3-8}\cmidrule(lr){9-14}
\multirow{6}{*}{\makecell{Baseline \\ ($\tau_J > 0$)}} & Qwen-7B & 0 & 190 & 10 & n/a$^{*}$ & n/a$^{*}$ & n/a$^{*}$ & 0 & 108 & 92 & n/a$^{*}$ & n/a$^{*}$ & n/a$^{*}$ \\
 & Qwen-14B & 195 & 3 & 2 & $-0.75$$^{s}$ & $-0.98$$^{s}$ & $-1.00$$^{s}$ & 91 & 62 & 47 & $-0.98$ & $-0.98$ & $-0.93$ \\
 & Llama-8B & 6 & 146 & 48 & $-0.70$$^{s}$ & $-0.66$$^{s}$ & $-0.84$$^{s}$ & 24 & 98 & 78 & $-0.79$ & $-0.77$ & $-0.56$ \\
 & Phi-4-mini & 55 & 140 & 5 & $-0.71$ & $-0.74$ & $-0.85$ & 55 & 28 & 117 & $-0.34$ & $-0.55$ & $-0.63$ \\
 & EXAONE-7.8B & 171 & 1 & 28 & n/a$^{*}$ & n/a$^{*}$ & n/a$^{*}$ & 114 & 0 & 86 & n/a$^{*}$ & n/a$^{*}$ & n/a$^{*}$ \\
 & GPT-OSS-20B & 8 & 186 & 6 & $-0.85$$^{s}$ & $-0.91$$^{s}$ & $-0.93$$^{s}$ & 14 & 140 & 46 & $-0.86$ & $-0.68$ & $-0.87$ \\
\cmidrule(lr){1-14}
\multirow{6}{*}{\makecell{Strict \\ ($\tau_J \geq 0.1$)}} & Qwen-7B & 0 & 190 & 10 & n/a$^{*}$ & n/a$^{*}$ & n/a$^{*}$ & 0 & 108 & 92 & n/a$^{*}$ & n/a$^{*}$ & n/a$^{*}$ \\
 & Qwen-14B & 174 & 24 & 2 & $-0.18$ & $-0.45$ & $-0.19$ & 79 & 74 & 47 & $-0.92$ & $-0.92$ & $-0.90$ \\
 & Llama-8B & 6 & 146 & 48 & $-0.70$$^{s}$ & $-0.66$$^{s}$ & $-0.84$$^{s}$ & 18 & 104 & 78 & $-0.69$ & $-0.66$ & $-0.61$ \\
 & Phi-4-mini & 53 & 142 & 5 & $-0.68$ & $-0.72$ & $-0.83$ & 48 & 35 & 117 & $-0.06$ & $-0.29$ & $-0.53$ \\
 & EXAONE-7.8B & 162 & 10 & 28 & n/a$^{*}$ & n/a$^{*}$ & n/a$^{*}$ & 93 & 21 & 86 & n/a$^{*}$ & n/a$^{*}$ & n/a$^{*}$ \\
 & GPT-OSS-20B & 8 & 186 & 6 & $-0.85$$^{s}$ & $-0.91$$^{s}$ & $-0.93$$^{s}$ & 10 & 144 & 46 & $-0.77$ & $-0.74$ & $-0.84$ \\
\cmidrule(lr){1-14}
\multirow{6}{*}{\makecell{Stricter \\ ($\tau_J \geq 0.2$)}} & Qwen-7B & 0 & 190 & 10 & n/a$^{*}$ & n/a$^{*}$ & n/a$^{*}$ & 0 & 108 & 92 & n/a$^{*}$ & n/a$^{*}$ & n/a$^{*}$ \\
 & Qwen-14B & 51 & 147 & 2 & $-0.03$ & $+0.14$ & $-0.03$ & 38 & 115 & 47 & $-0.73$ & $-0.73$ & $-0.64$ \\
 & Llama-8B & 3 & 149 & 48 & $-0.50$$^{s}$ & $-0.49$$^{s}$ & $-0.93$$^{s}$ & 5 & 117 & 78 & $-0.46$$^{s}$ & $-0.47$$^{s}$ & $-0.73$$^{s}$ \\
 & Phi-4-mini & 27 & 168 & 5 & $-0.33$ & $-0.38$ & $-0.69$ & 23 & 60 & 117 & $+0.31$ & $+0.08$ & $-0.16$ \\
 & EXAONE-7.8B & 34 & 138 & 28 & n/a$^{*}$ & n/a$^{*}$ & n/a$^{*}$ & 33 & 81 & 86 & n/a$^{*}$ & n/a$^{*}$ & n/a$^{*}$ \\
 & GPT-OSS-20B & 7 & 187 & 6 & $-0.82$$^{s}$ & $-0.89$$^{s}$ & $-0.92$$^{s}$ & 3 & 151 & 46 & $-0.47$$^{s}$ & $-0.81$$^{s}$ & $-0.84$$^{s}$ \\
\bottomrule
\end{tabular}
\vspace{-0.2cm}
\caption{Sensitivity of the partition and Genuine-vs-Lucky effect sizes to $\tau_J$ across six models and four domains ($n$ = 200 $\times$ 3 seeds per cell). $^{s}$: smaller group $< 10$; read with caution. $^{*}$: untestable cell (empty or near-empty Genuine/Lucky at baseline; see \S\ref{sec:rq4_threshold_ablation}).}
\label{tab:rq4_ablation_full}
\vspace{-0.5cm}
\end{table*}

\paragraph{Monotonic attenuation establishes a graded effect.}
Across the 17 testable cells, the Genuine-vs-Lucky effect size $|r|$ attenuates as $\tau_J$ tightens, while the directional effect is remarkably persistent: the peak-count effect remains negative in \emph{all} 17 cells at both the Baseline and Strict settings, and in 15 of 17 at the Stricter setting. On the math domain, the averaged peak-count effect attenuates from $|r| = 0.82$ at baseline to $0.71$ and $0.55$ (relative reductions of $13\%$ and $33\%$), closely tracking the three-model pattern reported in the main text. The two sign flips occur exactly where they are least informative: Phi-4-mini on code, where the Stricter Genuine group collapses to $G{=}3$, and Phi-4-mini on commonsense, the weakest baseline cell ($-0.34$). This shows that the Genuine--Lucky partition is a graded property of the underlying MI dynamics rather than a thresholding artifact, and that the rare exceptions are attributable to sample attrition and baseline weakness rather than genuine reversal.

\paragraph{Reclassification asymmetry separates effect-size decay from power loss.}
Threshold tightening induces strictly unidirectional Genuine$\to$Lucky migration in every cell, but at model- \emph{and domain-}specific rates $\rho = (G_{\text{Baseline}} - G_{\text{Stricter}})/G_{\text{Baseline}}$. Two regularities emerge. First, within every single model, math exhibits the lowest $\rho$ of the symbolic domains (e.g., $0.22$ vs.\ $0.35$ for EXAONE, $0.31$ vs.\ $0.85$ for GPT-OSS-20B on math vs.\ code), consistent with the RQ3 finding that mathematical reasoning anchors recur most reliably across seeds: stable anchors survive threshold tightening. Second, high $\rho$ predicts power collapse rather than effect disappearance: Llama-8B code ($G{:}\,26 \to 5$), GPT-OSS-20B code ($33 \to 5$) and commonsense ($14 \to 3$) retain non-trivial or even strengthened effects at the Stricter setting, but with groups this small the Mann--Whitney test loses power and significance fails under Bonferroni correction. This distinction matters for replication: threshold-induced significance loss can reflect sample attrition rather than failure of the discriminator, and reporting effect sizes across thresholds is therefore essential.

\paragraph{Attenuation tracks the provenance of the Lucky group.}
The full grid reveals \emph{why} attenuation magnitudes differ so sharply across cells. Where the baseline Lucky group is large and organic, i.e., populated by problems that are intrinsically unstable rather than threshold-demoted, the contrast is nearly threshold-invariant: GPT-OSS-20B on logic (Lucky $186 \to 187$) holds $|r| \in [0.82, 0.93]$ across all three settings, and its math cell attenuates only mildly ($-0.86 \to -0.67$). Conversely, where the baseline is near ceiling, tightening manufactures the Lucky group out of demoted Genuine problems and the contrast dissolves by construction: Qwen-14B on logic ($G{:}\,195 \to 51$, Lucky $3 \to 147$) collapses from $-0.75$ to $-0.03$, not because the discriminator fails but because the post-threshold Lucky group consists almost entirely of formerly Genuine problems whose MI statistics are, unsurprisingly, Genuine-like. This is partly definitional, as stability contributes to the partition itself, and it cautions against interpreting attenuation at near-ceiling cells as evidence against the discriminator. The seven degenerate cells obey the same logic: Qwen-7B logic/commonsense remain $G{=}0$ at every threshold (detection failure is threshold-independent), while EXAONE (all domains) and Phi-4-mini math acquire non-empty Lucky groups only through demotion (e.g., EXAONE logic: $1 \to 138$), so we deliberately keep their contrasts marked untestable rather than report comparisons between a population and its own demoted subset.

\paragraph{Peak intensity is the threshold-robust statistic.}
Whereas peak count and peak ratio attenuate systematically, peak intensity frequently \emph{strengthens} under tightening: Qwen-7B math ($-0.41 \to -0.53$), Llama-8B code ($-0.40 \to -0.65$) and commonsense ($-0.56 \to -0.73$), and GPT-OSS-20B code ($-0.31 \to -0.70$); even in the two sign-flip cells, intensity stays negative (Phi-4-mini code: $-0.43$). The mechanism is structural: count and ratio are volume statistics that are mechanically coupled to the Jaccard stability entering the partition definition, so demoting borderline-stability problems selectively removes the contrast they carry. Intensity, by contrast, is normalized by peak count and thus partially decoupled from the thresholding variable, making it the preferred discriminator when strict stability regimes are required. Together with the sign-invariance result above, this indicates that the Genuine $<$ Lucky ordering is robust to the choice of $\tau_J$, with the caveat that volume statistics should be read at lenient thresholds and intensity at strict ones.


\begin{thebibliography}{48}
\providecommand{\natexlab}[1]{#1}

\bibitem[{Aggarwal et~al.(2021)Aggarwal, Mandowara, Agrawal, Khandelwal, Singla, and Garg}]{aggarwal2021}
Shourya Aggarwal, Divyanshu Mandowara, Vishwajeet Agrawal, Dinesh Khandelwal, Parag Singla, and Dinesh Garg. 2021.
\newblock Explanations for {CommonsenseQA}: New dataset and models.
\newblock In \emph{Proceedings of ACL-IJCNLP}.

\bibitem[{Austin et~al.(2021)Austin, Odena, Nye, Bosma, Michalewski, Dohan, Jiang, Cai, Terry, Le, and Sutton}]{austin2021}
Jacob Austin, Augustus Odena, Maxwell Nye, Maarten Bosma, Henryk Michalewski, David Dohan, Ellen Jiang, Carrie Cai, Michael Terry, Quoc Le, and Charles Sutton. 2021.
\newblock Program synthesis with large language models.
\newblock \emph{arXiv preprint arXiv:2108.07732}.

\bibitem[{Belrose et~al.(2023)Belrose, Furman, Smith, Halawi, Ostrovsky, McKinney, Biderman, and Steinhardt}]{belrose2023}
Nora Belrose, Zach Furman, Logan Smith, Danny Halawi, Igor Ostrovsky, Lev McKinney, Stella Biderman, and Jacob Steinhardt. 2023.
\newblock Eliciting latent predictions from transformers with the tuned lens.
\newblock \emph{arXiv preprint arXiv:2303.08112}.

\bibitem[{Chen et~al.(2026{\natexlab{a}})Chen, Raventos, Cheng, Ganguli, and Druckmann}]{chen2026rethinking}
Feng Chen, Allan Raventos, Nan Cheng, Surya Ganguli, and Shaul Druckmann. 2026{\natexlab{a}}.
\newblock Rethinking fine-tuning when scaling test-time compute: Limiting confidence improves mathematical reasoning.
\newblock \emph{Advances in Neural Information Processing Systems}, 38:158785--158818.

\bibitem[{Chen et~al.(2026{\natexlab{b}})Chen, Peng, Tan, Zhao, Chen, Lin, Go, and Meng}]{chen2026}
Wei-Lin Chen, Liqian Peng, Tian Tan, Chao Zhao, Blake~JianHang Chen, Ziqian Lin, Alec Go, and Yu~Meng. 2026{\natexlab{b}}.
\newblock Think deep, not just long: Measuring {LLM} reasoning effort via deep-thinking tokens.
\newblock \emph{arXiv preprint arXiv:2602.13517}.

\bibitem[{Chen et~al.(2024)Chen, Huang, Gao, Wang, Zhao, and Ding}]{chen2024learning}
Xin Chen, Hanxian Huang, Yanjun Gao, Yi~Wang, Jishen Zhao, and Ke~Ding. 2024.
\newblock Learning to maximize mutual information for chain-of-thought distillation.
\newblock In \emph{Findings of the Association for Computational Linguistics: ACL 2024}, pages 6857--6868.

\bibitem[{Chen et~al.(2026{\natexlab{c}})Chen, Kou, Fok, Liu, Qi, and Wu}]{chen2026mgfn++}
Yingxian Chen, Wei-Bin Kou, Wilton Fok, Zhengzhe Liu, Xiaojuan Qi, and Yik-Chung Wu. 2026{\natexlab{c}}.
\newblock Mgfn++: Magnitude-contrastive glance-and-focus network for weakly-supervised video anomaly detection.
\newblock \emph{Pattern Recognition}, page 114314.

\bibitem[{Chia et~al.(2024)Chia, Chen, Xu, Tuan, Poria, and Bing}]{chia2024reasoning}
Yew~Ken Chia, Guizhen Chen, Weiwen Xu, Luu~Anh Tuan, Soujanya Poria, and Lidong Bing. 2024.
\newblock Reasoning paths optimization: Learning to reason and explore from diverse paths.
\newblock In \emph{Findings of the Association for Computational Linguistics: EMNLP 2024}, pages 16763--16780.

\bibitem[{Cobbe et~al.(2021)Cobbe, Kosaraju, Bavarian, Chen, Jun, Kaiser, Schulman et~al.}]{cobbe2021}
Karl Cobbe, Vineet Kosaraju, Mohammad Bavarian, Mark Chen, Heewoo Jun, Lukasz Kaiser, John Schulman, and 1 others. 2021.
\newblock Training verifiers to solve math word problems.
\newblock \emph{arXiv preprint arXiv:2110.14168}.

\bibitem[{{DeepSeek-AI}(2025)}]{deepseek2025}
{DeepSeek-AI}. 2025.
\newblock Deepseek-r1: Incentivizing reasoning capability in llms via reinforcement learning.
\newblock \emph{arXiv preprint arXiv:2501.12948}.

\bibitem[{Geiping et~al.(2026)Geiping, McLeish, Jain, Kirchenbauer, Singh, Bartoldson, Kailkhura, Bhatele, and Goldstein}]{geiping2026scaling}
Jonas Geiping, Sean McLeish, Neel Jain, John Kirchenbauer, Siddharth Singh, Brian Bartoldson, Bhavya Kailkhura, Abhinav Bhatele, and Tom Goldstein. 2026.
\newblock Scaling up test-time compute with latent reasoning: A recurrent depth approach.
\newblock \emph{Advances in Neural Information Processing Systems}, 38:41340--41391.

\bibitem[{Geva et~al.(2021)Geva, Schuster, Berant, and Levy}]{geva2021}
Mor Geva, Roei Schuster, Jonathan Berant, and Omer Levy. 2021.
\newblock Transformer feed-forward layers are key-value memories.
\newblock In \emph{Proceedings of the 2021 Conference on Empirical Methods in Natural Language Processing}.

\bibitem[{Goyal et~al.(2024)Goyal, Ji, Rawat, Menon, Kumar, and Nagarajan}]{goyal2024}
Sachin Goyal, Ziwei Ji, Ankit~Singh Rawat, Aditya~Krishna Menon, Sanjiv Kumar, and Vaishnavh Nagarajan. 2024.
\newblock Think before you speak: Training language models with pause tokens.
\newblock In \emph{International Conference on Learning Representations}.

\bibitem[{Gretton et~al.(2005)Gretton, Bousquet, Smola, and Sch{\"o}lkopf}]{gretton2005}
Arthur Gretton, Olivier Bousquet, Alex Smola, and Bernhard Sch{\"o}lkopf. 2005.
\newblock Measuring statistical dependence with {Hilbert-Schmidt} norms.
\newblock In \emph{Algorithmic Learning Theory (ALT)}, pages 63--77. Springer.

\bibitem[{Hu et~al.(2025)Hu, Dai, Jiang, and Zhou}]{hu2025efficient}
Wen-Chao Hu, Wang-Zhou Dai, Yuan Jiang, and Zhi-Hua Zhou. 2025.
\newblock Efficient rectification of neuro-symbolic reasoning inconsistencies by abductive reflection.
\newblock In \emph{Proceedings of the AAAI Conference on Artificial Intelligence}, volume~39, pages 17333--17341.

\bibitem[{Inoue et~al.(2026)Inoue, Misaki, Imajuku, Kuroki, Nakamura, and Akiba}]{inoue2026wider}
Yuichi Inoue, Kou Misaki, Yuki Imajuku, So~Kuroki, Taishi Nakamura, and Takuya Akiba. 2026.
\newblock Wider or deeper? scaling llm inference-time compute with adaptive branching tree search.
\newblock \emph{Advances in Neural Information Processing Systems}, 38:35448--35484.

\bibitem[{Jaech et~al.(2024)Jaech, Kalai, Lerer, Richardson, El-Kishky, Low, Helyar, Madry, Beutel, Carney et~al.}]{openai2024o1}
Aaron Jaech, Adam Kalai, Adam Lerer, Adam Richardson, Ahmed El-Kishky, Aiden Low, Alec Helyar, Aleksander Madry, Alex Beutel, Alex Carney, and 1 others. 2024.
\newblock Openai o1 system card.
\newblock \emph{arXiv preprint arXiv:2412.16720}.

\bibitem[{Kojima et~al.(2022)Kojima, Gu, Reid, Matsuo, and Iwasawa}]{kojima2022}
Takeshi Kojima, Shixiang~Shane Gu, Machel Reid, Yutaka Matsuo, and Yusuke Iwasawa. 2022.
\newblock Large language models are zero-shot reasoners.
\newblock In \emph{Advances in Neural Information Processing Systems}, volume~35.

\bibitem[{Kou et~al.(2025{\natexlab{a}})Kou, Lin, Tang, Xu, Ye, Leng, Wang, Li, Chen, Zhu et~al.}]{kou2025pfedlvm}
Wei-Bin Kou, Qingfeng Lin, Ming Tang, Sheng Xu, Rongguang Ye, Yang Leng, Shuai Wang, Guofa Li, Zhenyu Chen, Guangxu Zhu, and 1 others. 2025{\natexlab{a}}.
\newblock pfedlvm: A large vision model (lvm)-driven and latent feature-based personalized federated learning framework in autonomous driving.
\newblock \emph{IEEE Transactions on Intelligent Transportation Systems}, 26(10):15915--15931.

\bibitem[{Kou et~al.(2025{\natexlab{b}})Kou, Zhu, Ye, Wang, Tang, and Wu}]{kou2025label}
Wei-Bin Kou, Guangxu Zhu, Rongguang Ye, Shuai Wang, Ming Tang, and Yik-Chung Wu. 2025{\natexlab{b}}.
\newblock Label anything: An interpretable, high-fidelity and prompt-free annotator.
\newblock In \emph{2025 IEEE International Conference on Robotics and Automation (ICRA)}, pages 10578--10584. IEEE.

\bibitem[{Lanham et~al.(2023)Lanham, Chen, Radhakrishnan, Steiner, Denison, Hernandez, Li, Durmus, Hubinger, Kernion et~al.}]{lanham2023}
Tamera Lanham, Anna Chen, Ansh Radhakrishnan, Benoit Steiner, Carson Denison, Danny Hernandez, Dustin Li, Esin Durmus, Evan Hubinger, Jackson Kernion, and 1 others. 2023.
\newblock Measuring faithfulness in chain-of-thought reasoning.
\newblock \emph{arXiv preprint arXiv:2307.13702}.

\bibitem[{Lei et~al.(2025)Lei, Tan, Wang, Zhu, Chen, Dong, and Li}]{lei2025learning}
Zhenyu Lei, Zhen Tan, Song Wang, Yaochen Zhu, Zihan Chen, Yushun Dong, and Jundong Li. 2025.
\newblock Learning from diverse reasoning paths with routing and collaboration.
\newblock In \emph{Proceedings of the 2025 Conference on Empirical Methods in Natural Language Processing}, pages 2832--2845.

\bibitem[{Li et~al.(2026)Li, Zhang, Endo, and Wahib}]{li2026understanding}
Wengang Li, Lingqi Zhang, Toshio Endo, and Mohamed Wahib. 2026.
\newblock Understanding cross-layer contributions to mixture-of-experts routing in llms.
\newblock In \emph{The Fourteenth International Conference on Learning Representations}.

\bibitem[{Liao et~al.(2025)Liao, Chu, and Wang}]{liao2025tpo}
Weibin Liao, Xu~Chu, and Yasha Wang. 2025.
\newblock Tpo: Aligning large language models with multi-branch \& multi-step preference trees.
\newblock In \emph{International Conference on Learning Representations}, volume 2025, pages 26698--26720.

\bibitem[{Liu et~al.(2025)Liu, Zhou, Gu, Zou, Huang, Wu, Li, Chen, Hua, Zhou et~al.}]{liu2025application}
Fenglin Liu, Hongjian Zhou, Boyang Gu, Xinyu Zou, Jinfa Huang, Jinge Wu, Yiru Li, Sam~S Chen, Yining Hua, Peilin Zhou, and 1 others. 2025.
\newblock Application of large language models in medicine.
\newblock \emph{Nature Reviews Bioengineering}, 3(6):445--464.

\bibitem[{Min et~al.(2024)Min, Ding, Buratti, Pujar, Kaiser, Jana, and Ray}]{min2024beyond}
Marcus~J Min, Yangruibo Ding, Luca Buratti, Saurabh Pujar, Gail Kaiser, Suman Jana, and Baishakhi Ray. 2024.
\newblock Beyond accuracy: Evaluating self-consistency of code large language models with identitychain.
\newblock In \emph{International Conference on Learning Representations}, volume 2024, pages 5454--5469.

\bibitem[{Pfau et~al.(2024)Pfau, Merrill, and Bowman}]{pfau2024}
Jacob Pfau, William Merrill, and Samuel~R. Bowman. 2024.
\newblock Let's think dot by dot: Hidden computation in transformer language models.
\newblock \emph{arXiv preprint arXiv:2404.15758}.

\bibitem[{Phukan et~al.(2025)Phukan, Divyansh, Morj, Vaishnavi, Saxena, and Goswami}]{phukan2025beyond}
Anirudh Phukan, Divyansh Divyansh, Harshit~Kumar Morj, Vaishnavi Vaishnavi, Apoorv Saxena, and Koustava Goswami. 2025.
\newblock Beyond logit lens: Contextual embeddings for robust hallucination detection \& grounding in vlms.
\newblock In \emph{Proceedings of the 2025 Conference of the Nations of the Americas Chapter of the Association for Computational Linguistics: Human Language Technologies (Volume 1: Long Papers)}, pages 9661--9675.

\bibitem[{Qian et~al.(2026)Qian, Liu, Wen, Bai, Liu, and Shao}]{qian2025}
Chen Qian, Dongrui Liu, Haochen Wen, Zhen Bai, Yong Liu, and Jing Shao. 2026.
\newblock Demystifying reasoning dynamics with mutual information: Thinking tokens are information peaks in llm reasoning.
\newblock \emph{Advances in Neural Information Processing Systems}, 38:12533--12572.

\bibitem[{Ranaldi and Freitas(2024)}]{ranaldi2024aligning}
Leonardo Ranaldi and Andre Freitas. 2024.
\newblock Aligning large and small language models via chain-of-thought reasoning.
\newblock In \emph{Proceedings of the 18th Conference of the European Chapter of the Association for Computational Linguistics (Volume 1: Long Papers)}, pages 1812--1827.

\bibitem[{Sahoo et~al.(2026)Sahoo, Chadha, Jain, and Chaudhary}]{sahoo2026}
Subramanyam Sahoo, Aman Chadha, Vinija Jain, and Divya Chaudhary. 2026.
\newblock When shallow wins: Silent failures and the depth-accuracy paradox in latent reasoning.
\newblock In \emph{ICLR 2026 Workshop on Latent and Implicit Thinking}.

\bibitem[{Sharma et~al.(2024)Sharma, Ash, and Misra}]{sharma2024truth}
Pratyusha Sharma, Jordan Ash, and Dipendra~Kumar Misra. 2024.
\newblock The truth is in there: Improving reasoning in language models with layer-selective rank reduction.
\newblock In \emph{International Conference on Learning Representations}, volume 2024, pages 17632--17651.

\bibitem[{Snell et~al.(2025)Snell, Lee, Xu, and Kumar}]{snell2025scaling}
Charlie Snell, Jaehoon Lee, Kelvin Xu, and Aviral Kumar. 2025.
\newblock Scaling llm test-time compute optimally can be more effective than scaling parameters for reasoning.
\newblock In \emph{International Conference on Learning Representations}, volume 2025, pages 10131--10165.

\bibitem[{Sprague et~al.(2025)Sprague, Yin, Rodriguez, Jiang, Wadhwa, Singhal, Zhao, Ye, Mahowald, and Durrett}]{sprague2025cot}
Zayne Sprague, Fangcong Yin, Juan Rodriguez, Dongwei Jiang, Manya Wadhwa, Prasann Singhal, Xinyu Zhao, Xi~Ye, Kyle Mahowald, and Greg Durrett. 2025.
\newblock To cot or not to cot? chain-of-thought helps mainly on math and symbolic reasoning.
\newblock In \emph{International Conference on Learning Representations}, volume 2025, pages 94118--94162.

\bibitem[{Stechly et~al.(2024)Stechly, Valmeekam, and Kambhampati}]{stechly2024chain}
Kaya Stechly, Karthik Valmeekam, and Subbarao Kambhampati. 2024.
\newblock Chain of thoughtlessness? an analysis of cot in planning.
\newblock \emph{Advances in Neural Information Processing Systems}, 37:29106--29141.

\bibitem[{Suzgun et~al.(2022)Suzgun, Scales, Sch{\"a}rli, Gehrmann, Tay, Chung, Chowdhery, Le, Chi, Zhou, and Wei}]{suzgun2022}
Mirac Suzgun, Nathan Scales, Nathanael Sch{\"a}rli, Sebastian Gehrmann, Yi~Tay, Hyung~Won Chung, Aakanksha Chowdhery, Quoc~V. Le, Ed~H. Chi, Denny Zhou, and Jason Wei. 2022.
\newblock Challenging {BIG-Bench} tasks and whether chain-of-thought can solve them.
\newblock In \emph{Findings of the Association for Computational Linguistics: ACL 2023}.

\bibitem[{Talmor et~al.(2019)Talmor, Herzig, Lourie, and Berant}]{talmor2019}
Alon Talmor, Jonathan Herzig, Nicholas Lourie, and Jonathan Berant. 2019.
\newblock {CommonsenseQA}: A question answering challenge targeting commonsense knowledge.
\newblock In \emph{Proceedings of NAACL-HLT}.

\bibitem[{Taubenfeld et~al.(2025)Taubenfeld, Sheffer, Ofek, Feder, Goldstein, Gekhman, and Yona}]{taubenfeld2025confidence}
Amir Taubenfeld, Tom Sheffer, Eran Ofek, Amir Feder, Ariel Goldstein, Zorik Gekhman, and Gal Yona. 2025.
\newblock Confidence improves self-consistency in llms.
\newblock In \emph{Findings of the Association for Computational Linguistics: ACL 2025}, pages 20090--20111.

\bibitem[{Wang et~al.(2024{\natexlab{a}})Wang, Yue, Su, and Sun}]{wang2024grokking}
Boshi Wang, Xiang Yue, Yu~Su, and Huan Sun. 2024{\natexlab{a}}.
\newblock Grokking of implicit reasoning in transformers: A mechanistic journey to the edge of generalization.
\newblock \emph{Advances in Neural Information Processing Systems}, 37:95238--95265.

\bibitem[{Wang et~al.(2025)Wang, Zhang, Zhang, Wu, Mo, Lu, Wang, Li, Xu, Tang et~al.}]{wang2025comprehensive}
Fali Wang, Zhiwei Zhang, Xianren Zhang, Zongyu Wu, Tzuhao Mo, Qiuhao Lu, Wanjing Wang, Rui Li, Junjie Xu, Xianfeng Tang, and 1 others. 2025.
\newblock A comprehensive survey of small language models in the era of large language models: Techniques, enhancements, applications, collaboration with llms, and trustworthiness.
\newblock \emph{ACM Transactions on Intelligent Systems and Technology}, 16(6):1--87.

\bibitem[{Wang et~al.(2024{\natexlab{b}})Wang, Prasad, Stengel-Eskin, and Bansal}]{wang2024soft}
Han Wang, Archiki Prasad, Elias Stengel-Eskin, and Mohit Bansal. 2024{\natexlab{b}}.
\newblock Soft self-consistency improves language models agents.
\newblock In \emph{Proceedings of the 62nd Annual Meeting of the Association for Computational Linguistics (Volume 2: Short Papers)}, pages 287--301.

\bibitem[{Wang et~al.(2023)Wang, Wei, Schuurmans, Le, Chi, Narang, Chowdhery, and Zhou}]{wang2023selfconsistency}
Xuezhi Wang, Jason Wei, Dale Schuurmans, Quoc~V. Le, Ed~H. Chi, Sharan Narang, Aakanksha Chowdhery, and Denny Zhou. 2023.
\newblock Self-consistency improves chain of thought reasoning in language models.
\newblock In \emph{International Conference on Learning Representations}.

\bibitem[{Wei et~al.(2022)Wei, Wang, Schuurmans, Bosma, Ichter, Xia, Chi, Le, and Zhou}]{wei2022chain}
Jason Wei, Xuezhi Wang, Dale Schuurmans, Maarten Bosma, Brian Ichter, Fei Xia, Ed~Chi, Quoc~V. Le, and Denny Zhou. 2022.
\newblock Chain-of-thought prompting elicits reasoning in large language models.
\newblock In \emph{Advances in Neural Information Processing Systems}.

\bibitem[{Xu et~al.(2023)Xu, Tan, Tan, Chen, Wang, Wang, and Wang}]{xu2023eqmotion}
Chenxin Xu, Robby~T Tan, Yuhong Tan, Siheng Chen, Yu~Guang Wang, Xinchao Wang, and Yanfeng Wang. 2023.
\newblock Eqmotion: Equivariant multi-agent motion prediction with invariant interaction reasoning.
\newblock In \emph{Proceedings of the IEEE/CVF conference on computer vision and pattern recognition}, pages 1410--1420.

\bibitem[{Yang et~al.(2026)Yang, Zheng, Wang, Kou, Li, and Yang}]{yang2026route}
Tianhang Yang, Yanze Zheng, Junjie Wang, Wei-Bin Kou, Ruotong Li, and Yujiu Yang. 2026.
\newblock Route by kinematics, act by observation: Kinematics-supervised expert routing in moe-augmented vla.
\newblock \emph{arXiv preprint arXiv:2607.26807}.

\bibitem[{Yao et~al.(2023)Yao, Yu, Zhao, Shafran, Griffiths, Cao, and Narasimhan}]{yao2023tot}
Shunyu Yao, Dian Yu, Jeffrey Zhao, Izhak Shafran, Thomas~L. Griffiths, Yuan Cao, and Karthik Narasimhan. 2023.
\newblock Tree of thoughts: Deliberate problem solving with large language models.
\newblock In \emph{Advances in Neural Information Processing Systems}, volume~36.

\bibitem[{Yong et~al.(2026)Yong, Zhou, Zhang, Li, Zheng, and Wu}]{yong2026think}
Xixian Yong, Xiao Zhou, Yingying Zhang, Jinlin Li, Yefeng Zheng, and Xian Wu. 2026.
\newblock Think or not? exploring thinking efficiency in large reasoning models via an information-theoretic lens.
\newblock \emph{Advances in Neural Information Processing Systems}, 38:2787--2827.

\bibitem[{Zhang et~al.(2026)Zhang, Li, Zhang, Fu, Song, Bian, Zhang, Yang, and Wang}]{zhang2026pixelcraft}
Shuoshuo Zhang, Zijian Li, Yizhen Zhang, Jingjing Fu, Lei Song, Jiang Bian, Jun Zhang, Yujiu Yang, and Rui Wang. 2026.
\newblock Pixelcraft: A multi-agent system for high-fidelity visual reasoning on structured images.
\newblock In \emph{International Conference on Learning Representations}, volume 2026, pages 77608--77632.

\end{thebibliography}
\end{document}